\documentclass{article}

\usepackage[final]{neurips_2021}

\usepackage[utf8]{inputenc} 
\usepackage[T1]{fontenc}    
\usepackage{hyperref}       
\usepackage{url}            
\usepackage{booktabs}       
\usepackage{amsfonts}       
\usepackage{nicefrac}       
\usepackage{microtype}      

\usepackage{array}
\usepackage[table]{xcolor}

\usepackage[export]{adjustbox}
\usepackage[ruled,vlined]{algorithm2e}
\usepackage{amsthm}
\usepackage{amsmath}

\usepackage[labelfont=bf]{caption}

\title{Fast and Data-Efficient Training of Rainbow:\\an Experimental Study on Atari}

\author{%
  Dominik Schmidt 
  \\
  TU Wien\\
  \texttt{e11809917@student.tuwien.ac.at} \\
  \And
  Thomas Schmied \\
  TU Wien \\
  \texttt{e1553816@student.tuwien.ac.at} \\
}

\begin{document}

    \maketitle
    
    \begin{abstract}

            Across the Arcade Learning Environment, Rainbow achieves a level of performance competitive with humans and modern RL algorithms. However, attaining this level of performance requires large amounts of data and hardware resources, making research in this area computationally expensive and use in practical applications often infeasible. This paper's contribution is threefold: We (1) propose an improved version of Rainbow, seeking to drastically reduce Rainbow's data, training time, and compute requirements while maintaining its competitive performance; (2) we empirically demonstrate the effectiveness of our approach through experiments on the Arcade Learning Environment, and (3) we conduct a number of ablation studies to investigate the effect of the individual proposed modifications. Our improved version of Rainbow reaches a median human normalized score close to classic Rainbow's, while using 20 times less data and requiring only 7.5 hours of training time on a single GPU. We also provide our full implementation including pre-trained models.
            
    \end{abstract}
    
    \section{Introduction}

    In 2013, the Deep Q-Networks (DQN) algorithm \citep{mnih2013} kicked off a flurry of developments in the classic Atari reinforcement learning (RL) benchmark. In this benchmark, the agent is tasked with learning to play Atari 2600 games solely based on pixel inputs and rewards. While recent methods such as Agent57, R2D2, NGU, and MuZero \citep{agent57, r2d2, ngu, muzero} can now achieve above-human level performance in most or all games, a number of critical issues remain. Among these are the often vast data, time, and compute requirements for training these agents. Agent57, the only RL algorithm to achieve super-human performance on all 57 Atari games, required in the order of a century (78 billion frames) of gameplay experience in order to beat the last game, Atari Skiing \citep{agent57}.
    
    Rainbow \citep{rainbow}, introduced in 2017 and itself based on DQN, represents an important milestone in the development of the above-mentioned agents, acting as a foundation for Agent57 and other algorithms \citep{agent57, r2d2}. In the past, Rainbow has also served as a useful baseline for the development of new environments \citep{procgen, sonicretrocontest}. In this paper, we revisit the Rainbow algorithm and review some of the original design choices. By combining Rainbow with some more recent techniques and by simplifying the implementation, we seek to strike a good balance between performance, data efficiency, training time, and implementation complexity.
    
    
    Aside from the Atari benchmark, reinforcement learning has further achieved remarkable accomplishments in other domains, such as beating the human champions in some of the world's most difficult board games \citep{silver2016mastering, silver2017mastering, silver2018general}, as well as in strategic real-time combat video games \citep{vinyals2019grandmaster, berner2019dota}. Other prominent demonstrations of the effectiveness of RL are optimizing the energy consumption of data centers \citep{lazic2018data}, dexterous robotic manipulation of physical objects \citep{akkaya2019solving, andrychowicz2020learning}, and designing more efficient chip layouts \citep{mirhoseini2020chip} and neural architectures \citep{nasnet, pham2018efficient}.
    In these domains too, however, RL-based approaches can suffer from the issues of lacking data efficiency and requiring large amounts of compute resources, among others. Indeed, some of the most remarkable results achieved in these tasks required up to hundreds or even thousands of years worth of simulated environment interaction experience \citep{vinyals2019grandmaster, alphastarblog, berner2019dota, akkaya2019solving}. This severely limits the applicability of RL to many real-world tasks, where gathering such large amounts of experience is often infeasible, and the required hardware resources impose a substantial financial burden.
    
    One prominent stream of research that addresses these issues is data-efficient RL. This area refers to a set of algorithmic improvements to the RL agent that aim to enable it to learn from fewer environment interaction steps and shorten the training time the RL agent requires in order to learn the desired behavior. In recent years, this area has started to receive lots of attention within the RL community. Therefore, we discuss the literature on data-efficient RL in more detail in Section \ref{sec:relatedwork}.
    
    Other, more practical approaches attempt to explicitly shorten the time required to reach the learning objective by speeding up the RL training process itself. This can be achieved, for example, by employing simple tricks that are commonly used in other areas of machine learning but have previously rarely been used in RL. These include techniques to improve hardware utilization, mixed-precision training \citep{micikevicius2018mixed} and methods that aim to stabilize the training process.
    
    \subsection{Contributions}
    
    In this work, we integrate techniques from both streams of research into the established Rainbow agent while re-evaluating a number of Rainbow's original design choices. Among others, we investigate what effect a larger batch size, a bigger network, a higher learning rate, spectral normalization, and mixed precision have on the training speed and the RL agent's overall performance. While most of these tricks/enhancements have been studied individually in the context of RL, to the best of our knowledge, our work is the first to study them jointly. We discuss their appearance in the literature in more detail in Section \ref{sec:relatedwork}.
    
    To study these improvements, we rely on the well-established Atari benchmark. The Arcade Learning Environment (ALE) has long been an important challenge for RL research \citep{alebell, alebellrevis} and has particularly entered the spotlight since the introduction of Deep Q-Networks (DQN) \citep{mnih2013, mnih2015}. DQN combines Q-Learning with deep convolutional neural networks (CNNs) and experience replay to learn to play Atari games at human-level performance, solely from pixels and with no prior human knowledge. Since its inception, a variety of enhancements to the original DQN architecture have been proposed \citep{doubledqn, duelingdqn, per, rl_intro, noisydqn, distributional}, many of which have since been integrated into a single unified algorithm, Rainbow-DQN \citep{rainbow}. The introduction of Rainbow represents a critical milestone in the development of RL algorithms, and it has ever since been one of the best-performing and most established algorithms for Atari.
    
	Overall, we make the following contributions in this paper: 
		\begin{itemize}
	    \item We propose an improved version of Rainbow, seeking to reduce wall-clock training time and required hardware resources while improving data efficiency and maintaining competitive performance. 
	    \item We empirically demonstrate the effectiveness of our approach on the Atari benchmark, achieving a median human normalized score close to that of classic Rainbow-DQN while using 20 times less data. 
	    \item We conduct a number of ablation studies to investigate the effect of the proposed modifications individually and in aggregate.
	\end{itemize}
	
	\newpage

    \section{Related Work} \label{sec:relatedwork}
    This section covers some recent directions of research in deep reinforcement learning that are orthogonal to our work.
    \subsection{Data-efficient RL: the algorithmic solution}
    One prominent stream of research related to our work is data-efficient RL. Data-efficient RL aims to learn from fewer environment interaction steps by leveraging the information at the agent's disposal more effectively. Recently, this has become an attractive area of research, as current RL algorithms can be incredibly data-inefficient. In fact, data inefficiency has been identified as one of the major obstacles towards widespread adoption of RL in the real world \citep{dulac2019challenges, dulac2020empirical}. Different solutions have been proposed to mitigate this limitation. Three of the most promising ones are (1) self-supervised learning, (2) model-based RL, and (3) data augmentation. 
    
    \noindent \textbf{Self-supervised learning.} Self-supervised methods in RL aim to learn more effective representations of the environment by optimizing a self-supervised auxiliary objective in addition to the primary RL objective. Self-supervised learning (SSL) is the driving force behind recent advances in NLP \citep{devlin2018bert, gpt3}, computer vision \citep{chen2020big, grill2020bootstrap} and speech recognition \citep{baevski2020wav2vec}. Only recently, SSL has started to receive widespread attention from the RL community and shown to improve data efficiency considerably. Prominent examples are Unreal \citep{unreal}, CURL \citep{curl}, SPR \citep{schwarzer2020data}, and SGI \citep{schwarzer2021pretraining}.
    
    \noindent \textbf{Model-based methods.} Model-based RL algorithms, in contrast, learn an explicit model of the environment and then use this model to derive a good policy by simulating experience, a process known as planning \citep{rl_intro}. A major advantage, again, is more data-efficient learning, but these methods have historically proven difficult in environments with large and complex observation spaces \citep{mbasedrl}. However, \citet{muzero} recently introduced MuZero, a model-based RL algorithm that achieves state-of-the-art performance on the Arcade Learning Environment \citep{alebell}. 
    Other prominent examples of model-based RL methods include World Models \citep{worldmodels}, SimPle \citep{kaiser2019model} and Dreamer \citep{hafner2019dream, hafner2020mastering}. A detailed overview of the model-based RL literature is given by \citet{mbasedrl}.
    
    \noindent \textbf{Data augmentation.} Another option to improve data efficiency in RL is data augmentation, a well-established technique in many ML domains, most notably for image-based tasks \citep{perez2017effectiveness, shorten2019survey}. Data augmentation aims to artificially increase the amount of available training data by modifying existing observations in some meaningful way. For RL, this simple technique has been shown to improve data efficiency considerably, as demonstrated by RAD \citep{laskin2020reinforcement} and DrQ \citep{yarats2020image}. 
    
    \subsection{Speeding up RL training: the practical solution} \label{sec:practicalspeedup}
    Both the approaches discussed in this section, as well as the data-efficient methods from the previous section, deal with the same fundamental question: how to accelerate the training of reinforcement learning agents. On the one hand, data-efficient methods implicitly shorten the training time by cutting down the number of required training samples. The methods discussed in this section, on the other hand, seek to shorten the required training time without fundamentally modifying the learning algorithm. These methods are not unique to deep RL but rather are generally applicable to many deep learning algorithms. We do not cover approaches that speed up training through parallelization since these simply trade hardware requirements for training time and thus do not alleviate the financial burden of training large models.
    
    \noindent \textbf{Larger batch sizes.} For image-based RL tasks, agents are typically trained with relatively small batch sizes \citep{mnih2015, rainbow}. However, prior work has demonstrated that, on the procgen and ALE benchmarks, increasing the batch size can speed up the training time considerably \citep{procgen, acceleratedrl}.
    
    \noindent \textbf{Bigger networks.} Compared to other branches of deep learning, the networks employed by deep RL agents are relatively shallow \citep{mnih2015}. Replacing the standard network architecture with bigger networks has proven a fruitful direction in prior work \citep{impala, procgen, sinha2020d2rl, bjorck2021towards}.
    
    \noindent \textbf{Spectral normalization.} Recently, \citet{specnorm} proposed to apply Spectral Normalization (SN), a technique that originates from the literature on Generative Adversarial Networks (GANs) \citep{miyato2018spectral, kurach2019large}, to the RL setting. The authors observed that a Categorical DQN agent (C51) \citep{distributional} augmented with SN achieves similar results as a full Rainbow DQN implementation \citep{rainbow}.
    
    \noindent \textbf{Mixed precision.} Another prominent and well-established option to accelerate the training of neural networks is mixed precision, initially proposed by \citet{micikevicius2018mixed}. In RL, \citet{lam2019quantized} and \citet{lowprecrl} were the first to integrate low/mixed precision and saw substantial improvements in training time.
    
    \subsection{RL \& the Atari benchmark: a lasting relationship}
    The Arcade Learning Environment (ALE) \citep{alebell, alebellrevis} is a lasting and indispensable element of the RL researcher's toolbox. It is also the focus of our work. Since its inception, hundreds of RL algorithms have been developed, and the achieved scores have only increased over time. The first model-free architecture to raise public interest on the Atari benchmark was Deep Q-Networks (DQN) \citep{mnih2013, mnih2015}. Several enhancements to DQN have been proposed.
    including Double DQN \citep{doubledqn}, Dueling DQN \citep{duelingdqn}, Prioritized Experience Replay \citep{per}, Multi-step Bootstrapping \citep{rl_intro}, Noisy Nets \citep{noisydqn}, and Distributional RL \citep{distributional}.
    
    Rainbow DQN, the architecture we leverage in this work, integrates all these enhancements into a single unified framework and shows that they are largely compatible \citep{rainbow}. A slightly modified version of Rainbow introduced by \citet{der} and referred to as Data-efficient Rainbow trades off reduced computational efficiency in exchange for significantly improved data efficiency.
    
    \noindent R2D2 \citep{r2d2} and Agent57 \citep{agent57}, distributed RL algorithms, follow in the lineage of Rainbow. Similar distributed training strategies were employed by Gorila \citep{gorila}, Ape-X \citep{apex}, IMPALA \citep{impala} and Reactor \citep{reactor}. Like IMPALA, R2D2 also uses recurrent neural networks to take advantage of longer state histories. Agent57 builds off of the methods developed in \citet{r2d2} but combines them with two intrinsic reward-based exploration mechanisms: the curiosity-based Random Network Distillation \citep{rnd} and Never Give Up exploration \citep{ngu}, aimed at increasing long and short term state-space coverage, respectively. Additionally, a UCB bandit \citep{rl_intro} algorithm is used to control the amount of exploration and the value of the discount factor in each of the distributed actors. 
    
    
    \section{Preliminaries}
    In this section, we briefly discuss the theoretical foundations of this work. 
    \subsection{The RL framework} 
    \textbf{Markov decision process.} The fundamental goal of RL is to learn how to behave in an unknown environment in order to maximize a scalar reward signal. In each step of this sequential decision-making task, the agent observes the environment's current state $s_t$, performs some action $a_t$, transitions to the next state $s_t$, and obtains a reward $r_t$ \citep{rl_intro}. We assume the standard formulation of the Markov Decision Process (MDP) with $\langle \mathcal{S}, \mathcal{A}, R, P \rangle$ where
    \begin{itemize}
        \item $\mathcal{S}, \mathcal{A}$ are state and action space, respectively
        \item $R\colon \mathcal{S} \times \mathcal{A} \times \mathcal{S} \to \mathbb{R}$ is the reward function that assigns each transition from state $s_t$ to state $s_{t+1}$ via action $a_t$ a real valued reward $r_t = R(s_t, a_t, s_{t+1})$
        \item $P\colon \mathcal{S} \times \mathcal{A} \to [0, 1]$ is the transition probability function that specifies a conditional probability $p(s_{t+1} | s_t, a_t)$ of transitioning into state $s_{t+1}$ after executing action $a_t$ in state $s_t$.
    \end{itemize}
    \textbf{The Goal of RL.} The goal the agents pursues is to maximize its return, $G_t = \sum_{k=t+1}^T R_k$, the sum of all collected rewards. To achieve this, it learns a policy $\pi$ that determines its behavior. The policy $\pi$ maps the perceived states to actions $a \sim \pi(a \mid s) $. The policy $\pi$ can be either learned directly via the policy gradient or indirectly via the Bellman equations. RL algorithms that directly learn the policy are referred to as policy-based methods whereas algorithms that learn value functions are known as value-based methods \citep{rl_intro}. 

    \textbf{Bellman equations.} The state-value function $v_\pi(s) = \mathbb{E}_{a \sim \pi}[r(s, a) + \gamma v_\pi(s')]$ and action value function $q_\pi(s, a) = \mathbb{E}[r(s, a) + \gamma \mathbb{E}_{a' \sim \pi}[q_\pi(s', a')]]$ determine how good it is to be in a certain state and how good it is to perform a certain given a certain state, respectively \citep{rl_intro}. 

	\subsection{Q-learning: from tables to neural networks}
	\textbf{Tabular Q-learning} Q-Learning is a value-based RL algorithm that explicitly maintains a table of Q-estimates for state-action pairs \citep{watkins1992q}. More generally, it falls under temporal difference (TD) learning methods as it grounds value estimates for the current time step in estimates at future steps. The Q-estimates are repeatedly updated according to: $q(s, a) \gets q(s, a) + \alpha [r + \gamma \max_a q(s', a)- q(s, a)]$. While Q-Learning works well for small problems, storing the table becomes intractable for larger state/action spaces \citep{rl_intro}. 

    \textbf{Deep Q-Network.} DQN, proposed by \citet{mnih2015}, employ neural networks to represent the Q-table. In contrast to tabular Q-learning, DQN scales large state and action spaces. Furthermore, \citet{mnih2015} introduced two additional enhancements to Q-Learning that are essential for good performance when using function approximation, experience replay as well as a separate target network that is updated periodically. Therefore, the learning objective becomes:  
    \begin{equation}\label{eq:dqn}
    J(\theta) = \mathbb{E}_{(s, a, s', r) \sim \mathcal{D}} [( r + \gamma \max_{a'} q_{\theta^-}(s, a') - q_{\theta}(s, a))^2 ]
\end{equation}
    where $\mathcal{D}$ represents the experience replay buffer, and $\theta$ and $\theta^-$ are the neural networks that parameterize the Q-functions.

	\section{Our Approach}
    
    Our main goals in this work are to reduce Rainbow's considerable hardware, training time, and data requirements while maintaining a similar level of performance as the original agent. We achieve these goals by combining a large Q-network architecture with an efficient high-throughput implementation as well as by accelerating and stabilizing training through extensive hyper-parameter tuning and the use of spectral normalization. Furthermore, we reduce the implementation complexity of Rainbow by removing the distributional RL component. While distributional RL was essential for good performance in \citet{rainbow} when training for over 40M frames, our improvements in learning speed --- leading to lower overall required training time --- made distributional RL less vital.
    
    Our implementation is available at \url{https://github.com/schmidtdominik/Rainbow} and includes a complete and highly customizable framework for preprocessing, training, and evaluation. We further provide integrations for OpenAI's \texttt{gym}, \texttt{procgen}, and \texttt{gym-retro} environments, as well as pretrained models for all 53 tested Atari games. 
    
    In the rest of this section, we discuss the tricks we apply and describe the individual components of our approach. First, we describe our evaluation methodology. Then, we describe the network architecture we employ in this work and show how different variants thereof compare. Furthermore, we discuss how we make use of spectral normalization and again compare a few variants. Also, we address the tricks we employ to improve the hardware utilization of our agent. 
    
    \subsection{Evaluation methodology}
	We evaluate our approach against the same set of 54 Atari games that were used in \citet{rainbow}, excluding the game \emph{Surround} as it is not available via OpenAI \texttt{gym}.
	To this end, we closely follow the evaluation procedure from \citet{mnih2015} and \citet{rainbow}. All evaluation runs lasted for 500k frames and each individual episode was no longer than 108k frames.
	The only modification we made was that we performed the evaluation runs after training had concluded by periodically saving model snapshots during training and later loading them for evaluation. Lastly, we re-evaluated the best-performing snapshot for each game. Each experiment was performed with three random seeds.
	
	Due to the high computational cost of training on the whole ALE, we limited our ablation studies to a subset of 5 games (Asterix, Beam Rider, Freeway, Seaquest, and Space Invaders) as recommended in \citet{alebellrevis} and \citet{alebell}. This selection includes two human-optimal games, two score-exploitable games, and one sparse reward game \citep{alegamedesc}. Overall, this selection intends to capture the large variety of the ALE in a small number of games.

	We encountered one issue with comparability that occurs since \citet{rainbow} trained their agent for 200M frames while we only trained for 10M frames: the frequency of evaluation snapshots is the same (every 1M frames), but the total number is not. This means that for some games where the performance fluctuates wildly during training, taking a larger number of agent snapshots could increase the likelihood of taking at least one snapshot at a point of good performance. To support this hypothesis, we separately stored and evaluated twice the number of agent snapshots (every 500k training frames) and observed an approximately 10\% increase in apparent performance. This suggests that evaluation results from agents trained on the ALE and evaluated on a lower number of snapshots than previous research may underestimate the true performance of their agent.
    
	\subsection{Larger and deeper Q-Network Architecture}
	First, we replace the small dueling network architecture, as introduced in \citet{mnih2013} and combined with dueling DQN in \citet{duelingdqn}, with the both larger and deeper IMPALA CNN \citep{impala}. More specifically, we employ the large variant of the IMPALA CNN (see Figure 1 in the supplementary material) with twice the number of channels, as modified in \citet{procgen}. We additionally add a size $6 \times 6$ adaptive max-pooling layer between the network's convolutional and fully connected parts. This simple yet powerful modification makes it straightforward to use our implementation with inputs of different resolutions (such as games provided by \texttt{procgen} or \texttt{gym-retro}) without affecting the number of parameters in the network.
	Like \citet{procgen} and \citet{impala}, we found that using this architecture substantially increased learning speed (both in terms of wall-clock time and training steps), sample efficiency, and final overall performance.
		\begin{figure}[!h]
    	\renewcommand{\arraystretch}{8}
    	\setlength\tabcolsep{1pt}
        \centering
    	\begin{tabular}{lll}
    	\includegraphics[width=.33\linewidth,valign=m]{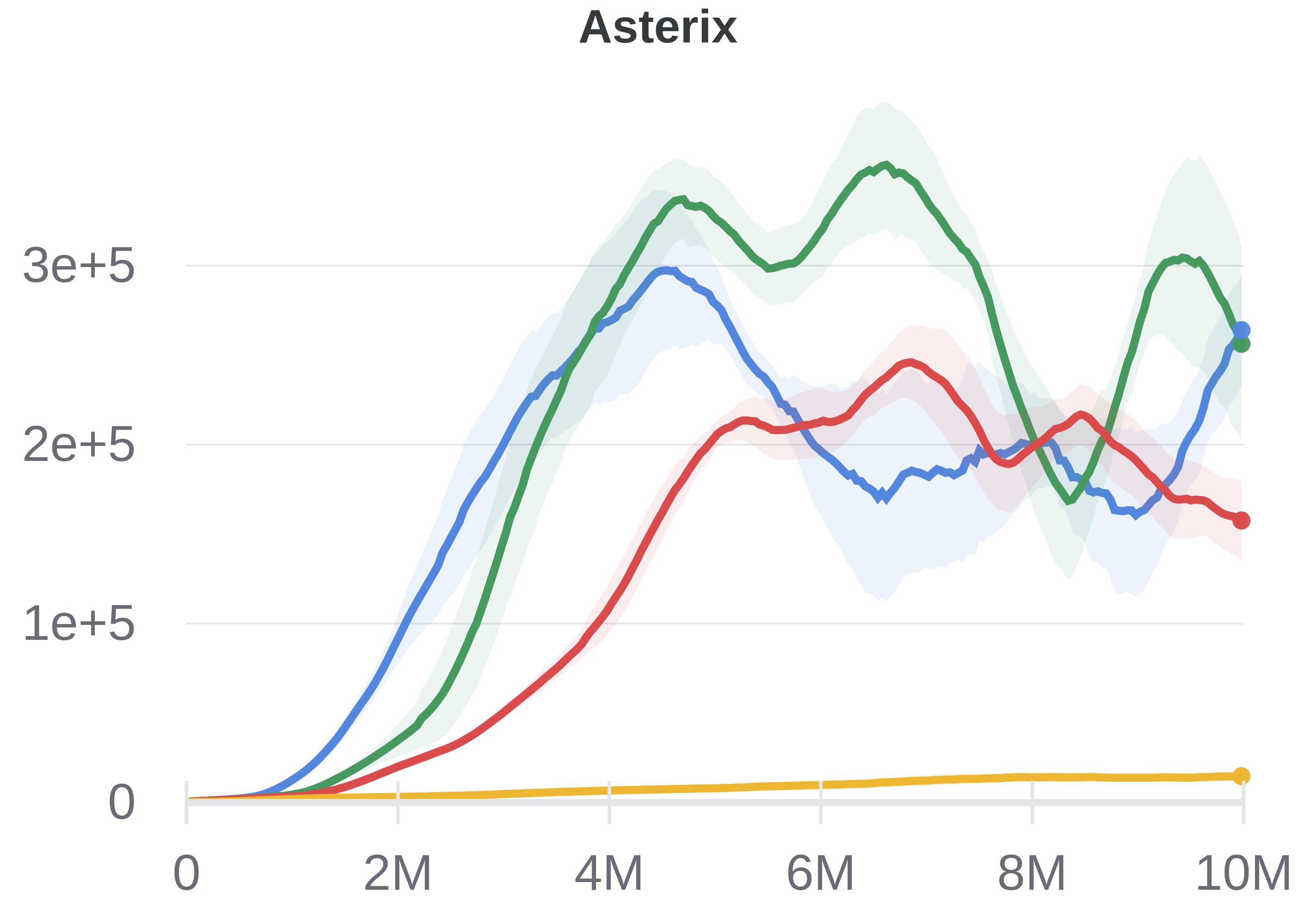} & 
    	\includegraphics[width=.33\linewidth,valign=m]{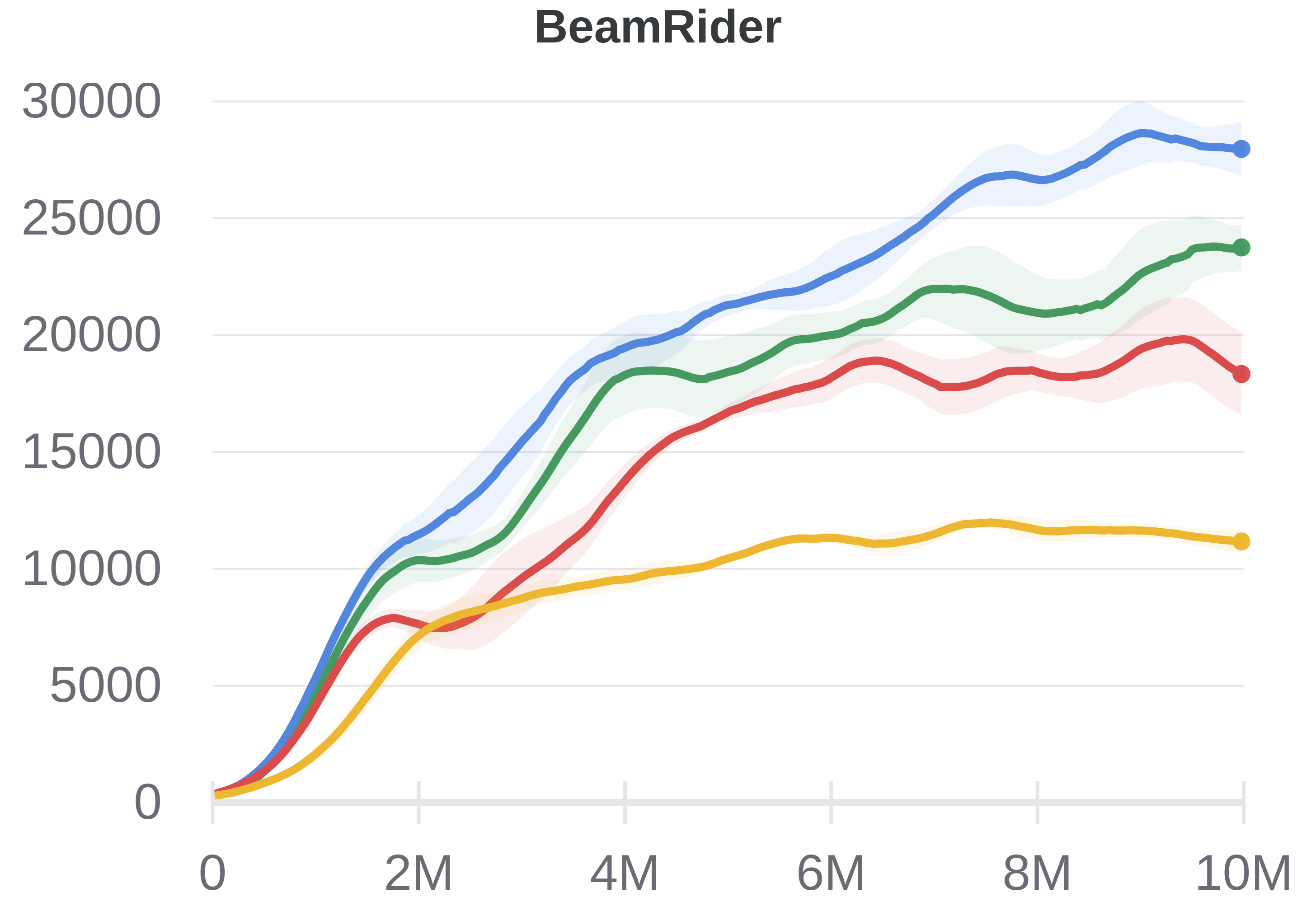} & 
    	\includegraphics[width=.33\linewidth,valign=m]{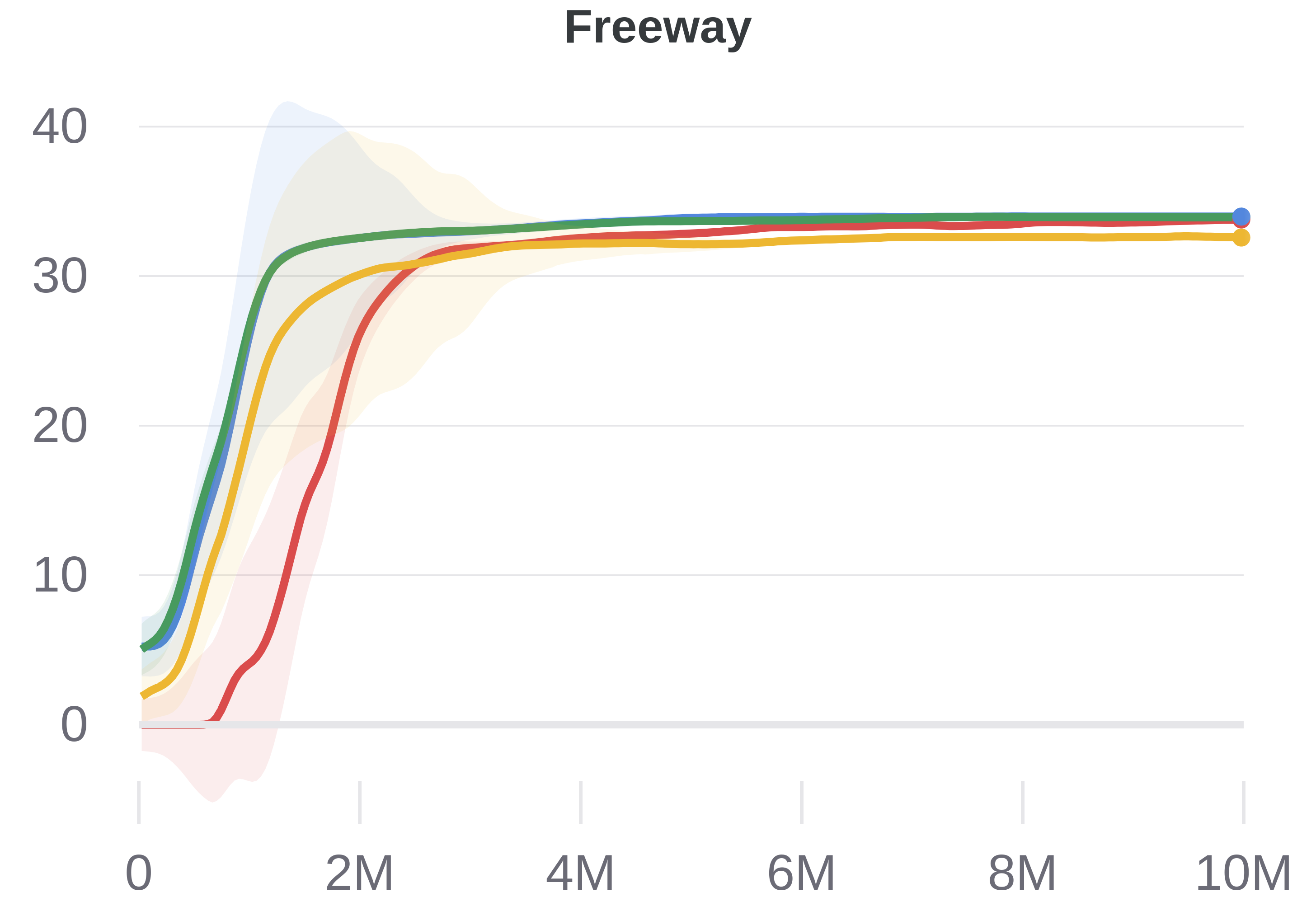} \\
    	\includegraphics[width=.33\linewidth,valign=m]{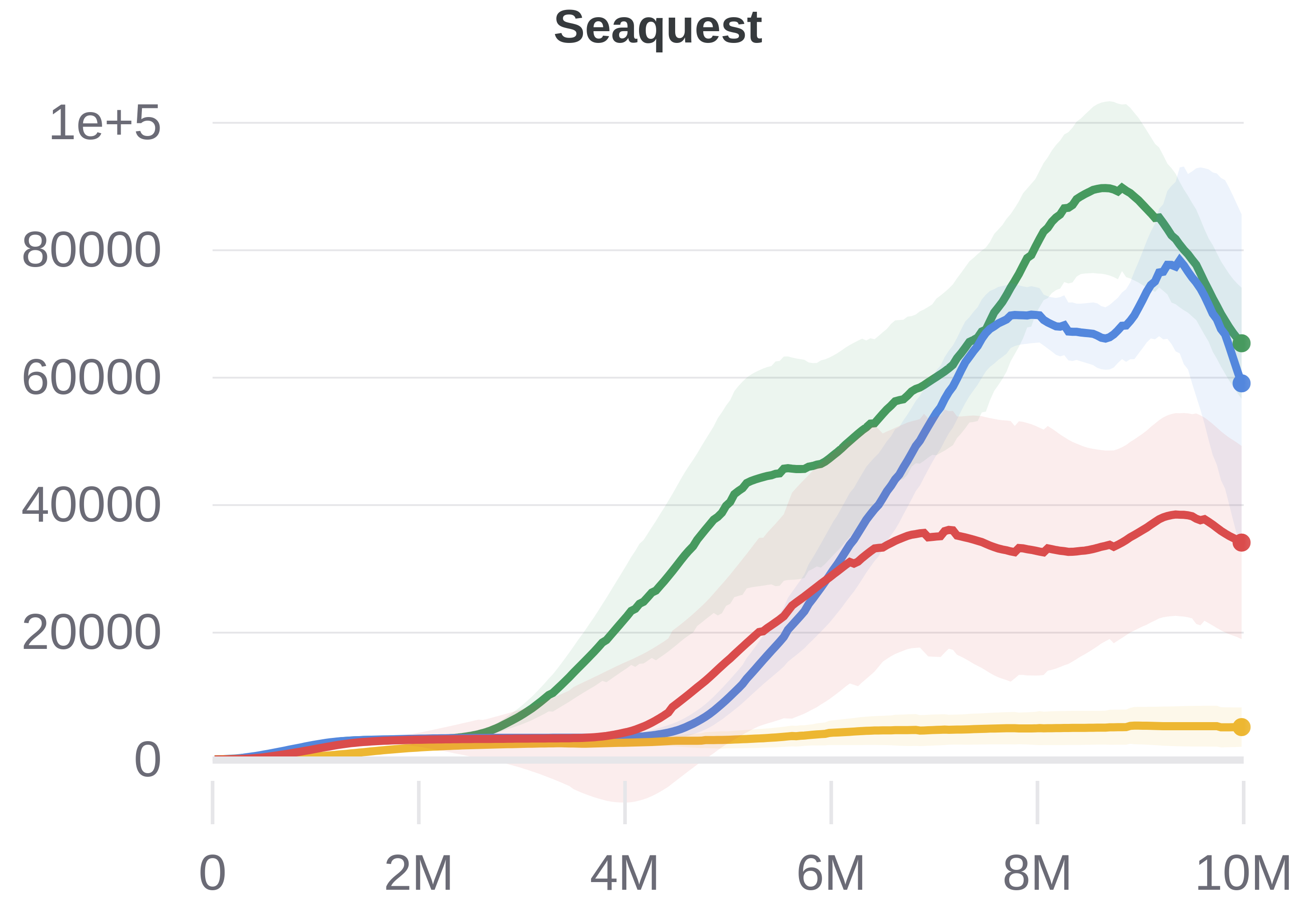} & 
    	\includegraphics[width=.33\linewidth,valign=m]{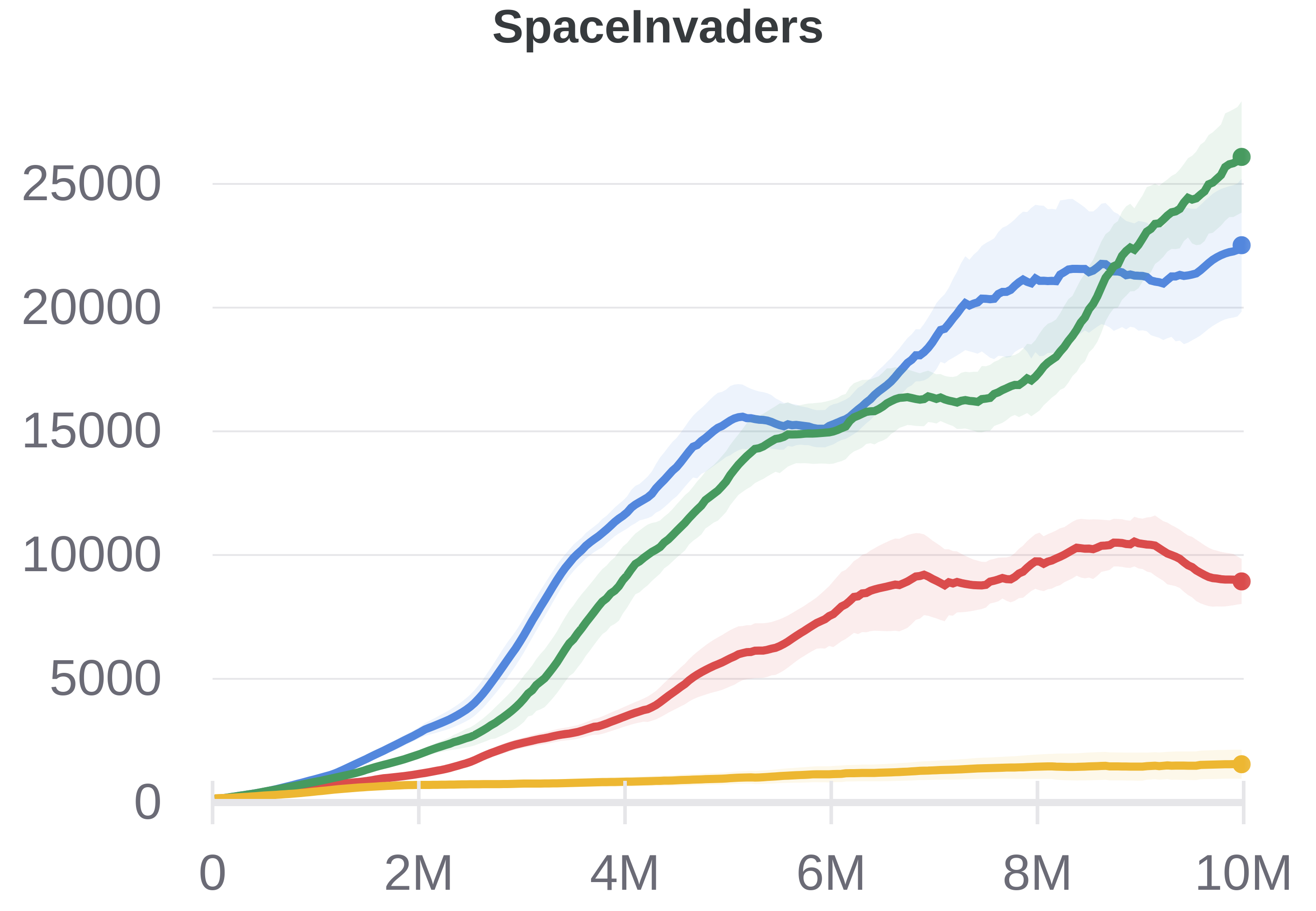} & 
    	\includegraphics[width=.33\linewidth,valign=m]{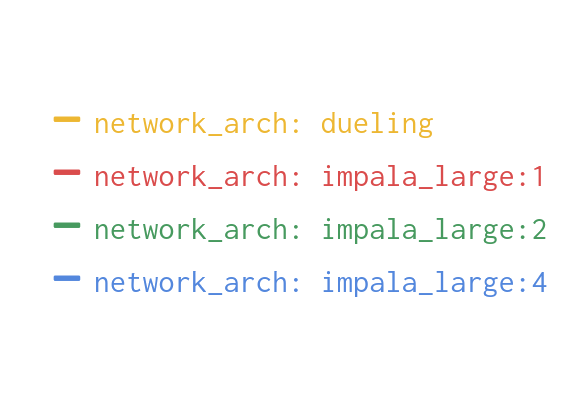}
    	\end{tabular}
    	\vspace*{7pt}
    	\caption{Comparison of the different network architectures. \emph{Dueling} refers to the "Nature" dueling architecture used in \citet{rainbow}. Each curve shows the median over three seeds of the 100-episode running average of episode returns as a function of environment interactions.}
    	\label{fig:impalanet}
    \end{figure}
    
	In our experiments, we compared values of 1, 2 and 4 for the channel multiplier. The results of these experiments are shown in Figure \ref{fig:impalanet}. We found that a value of 2 yielded a good trade-off between data and computational efficiency. Consequently, we selected this network architecture for our subsequent runs. 
	
	\subsection{Spectral Normalization}

	Spectral normalization (SN), commonly used to stabilize the training of discriminators in GANs \citep{miyato2018spectral}, is a method for controlling the Lipschitz constant of linear operators such as convolutional or dense linear layers in neural networks. Spectral normalization can be efficiently approximated via the power iteration method \citep{miyato2018spectral, specnorm}.
	
	In our experiments, we compared three different variants of applying spectral normalization to the convolutional layers in the IMPALA CNN:
	\begin{itemize}
	    \item \emph{none} -- no spectral normalization is performed.
	    \item \emph{all} -- SN is applied to both convolutional layers in all six residual blocks.
	    \item \emph{last} -- SN is applied to all convolutional layers in  the final two residual blocks.
	\end{itemize}
	
	\begin{figure}[!h]
        \renewcommand{\arraystretch}{8}
    	\setlength\tabcolsep{1pt}
        \centering
    	\begin{tabular}{lll}
    	\includegraphics[width=.33\linewidth,valign=m]{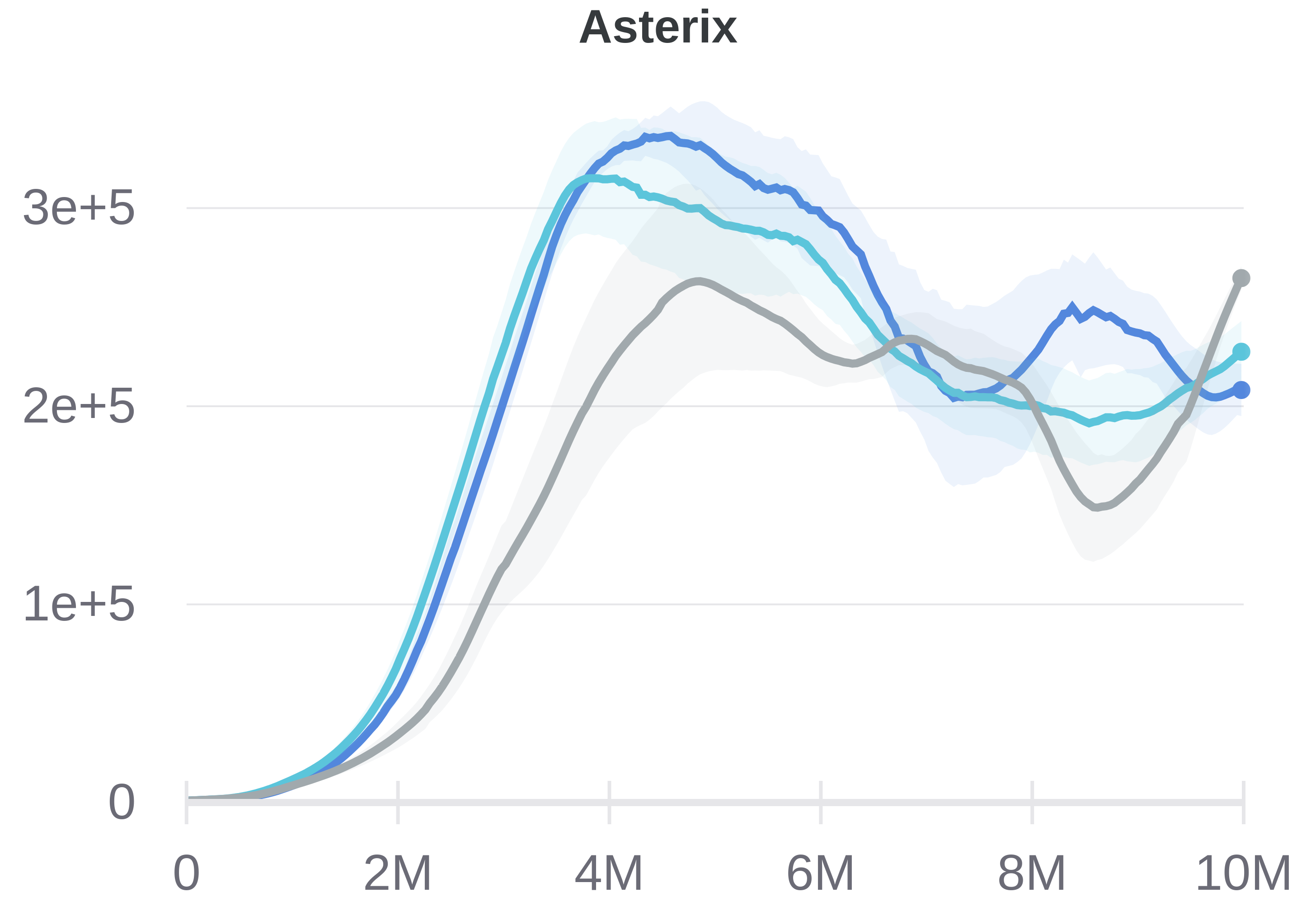} & 
    	\includegraphics[width=.33\linewidth,valign=m]{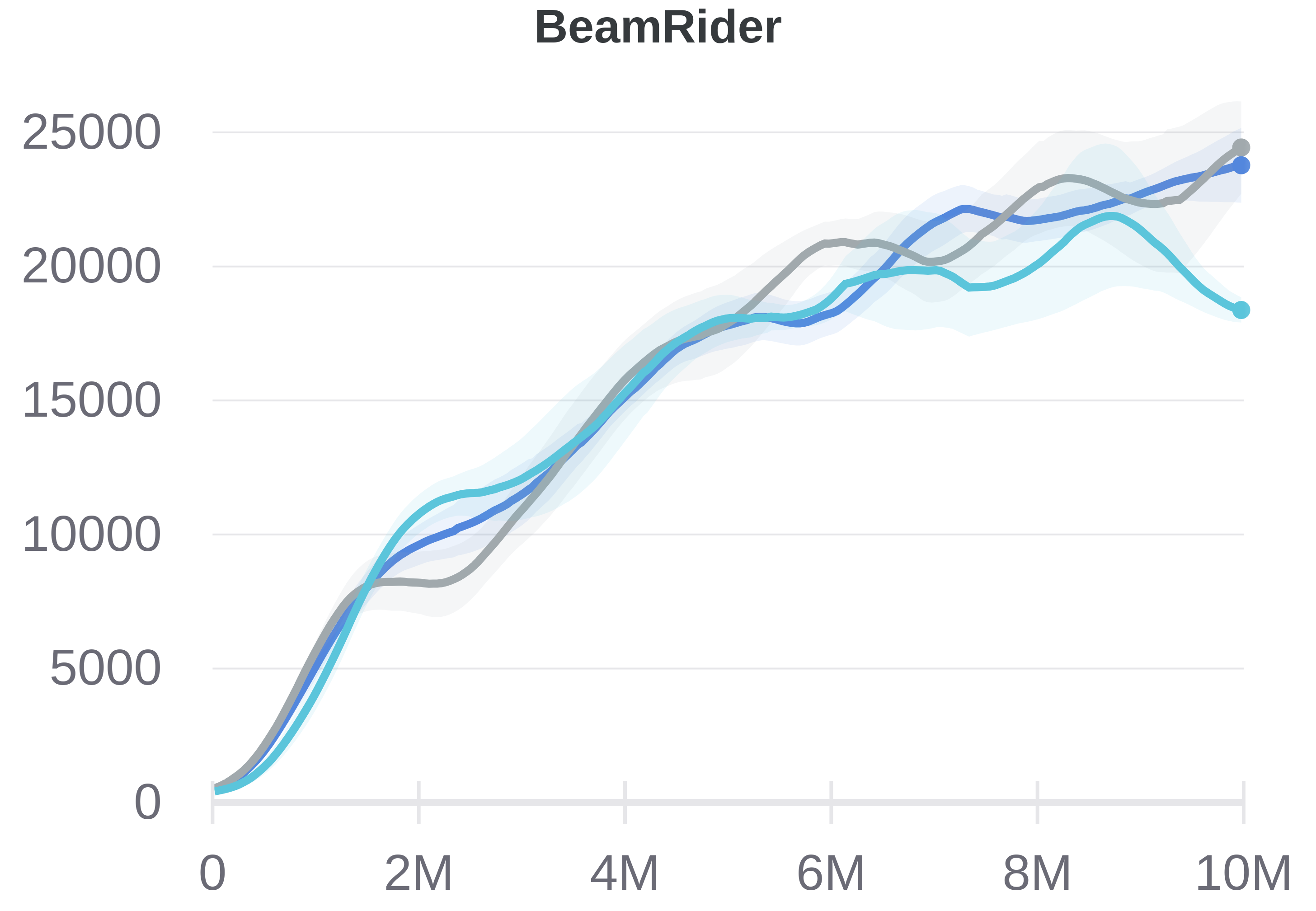} & 
    	\includegraphics[width=.33\linewidth,valign=m]{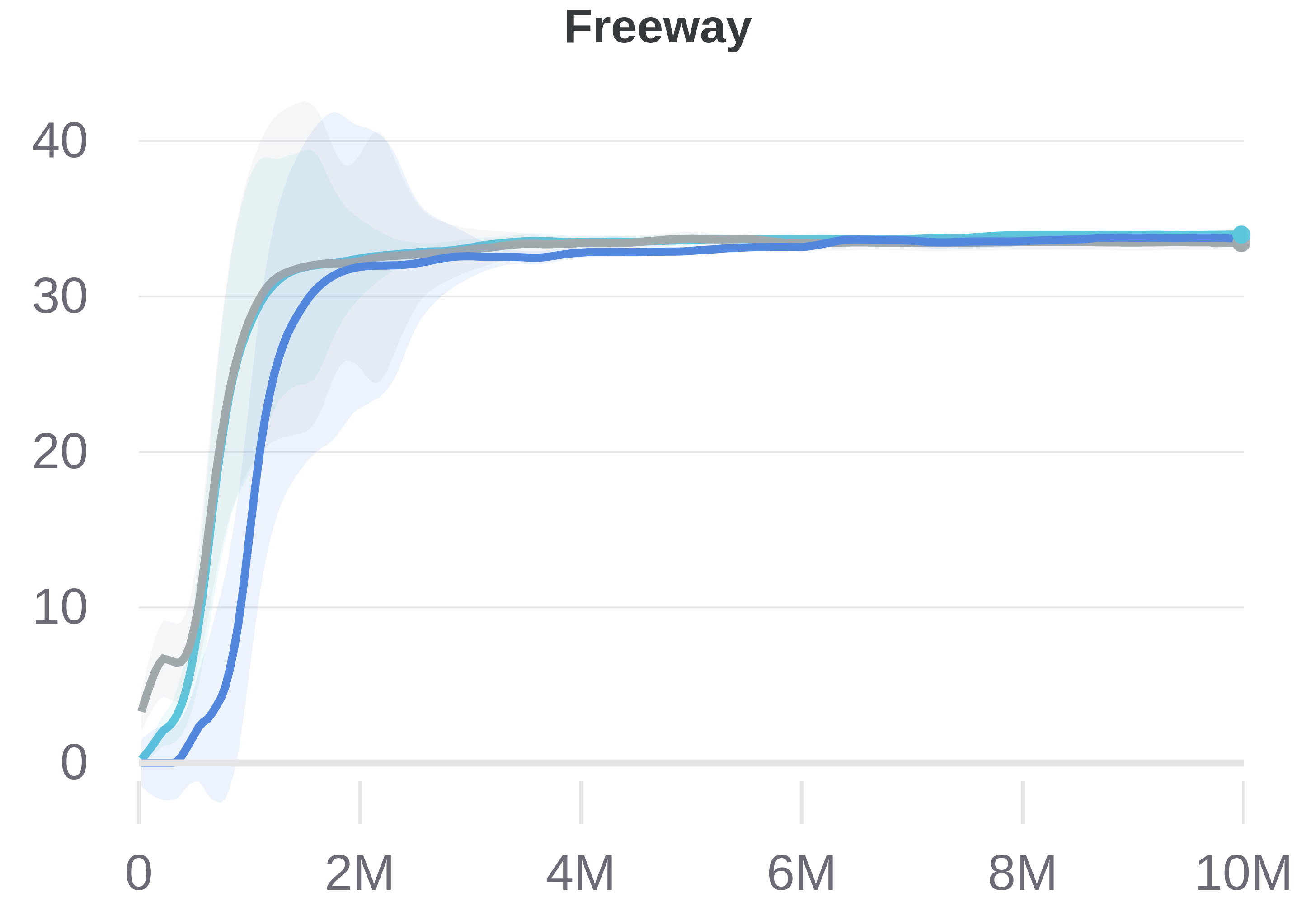} \\
    	\includegraphics[width=.33\linewidth,valign=m]{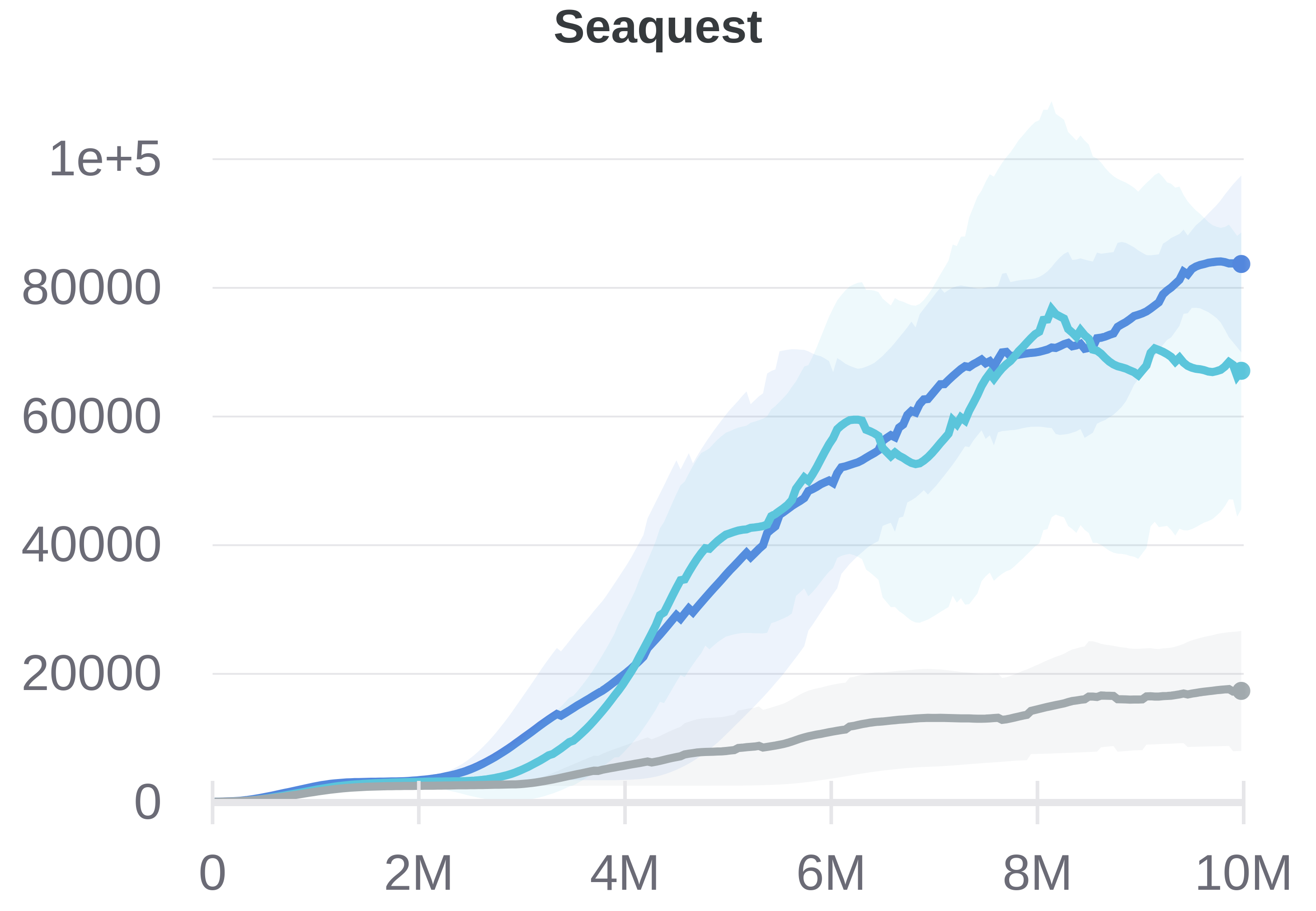} &
    	\includegraphics[width=.33\linewidth,valign=m]{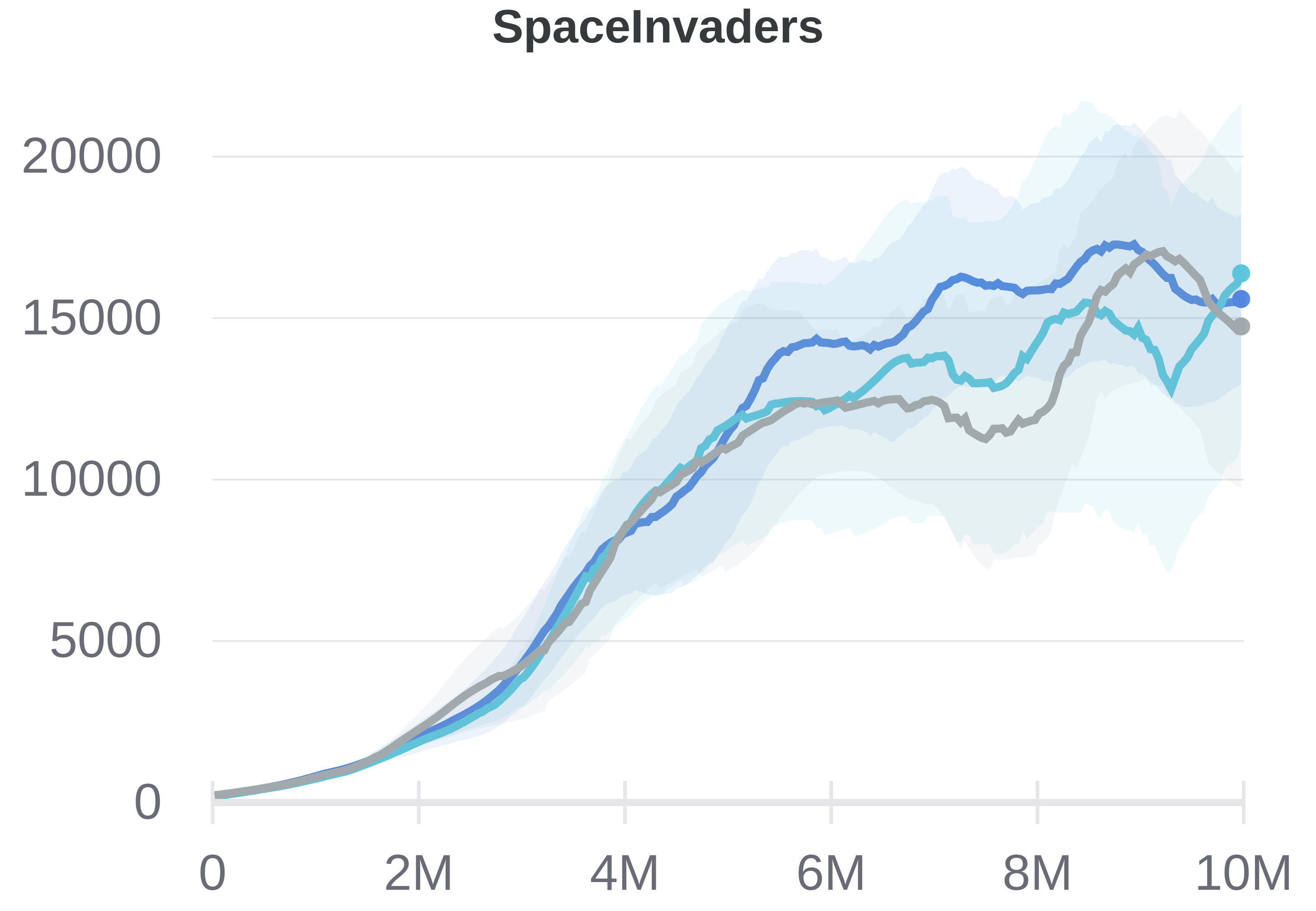} & 
    	\includegraphics[width=.33\linewidth,valign=m]{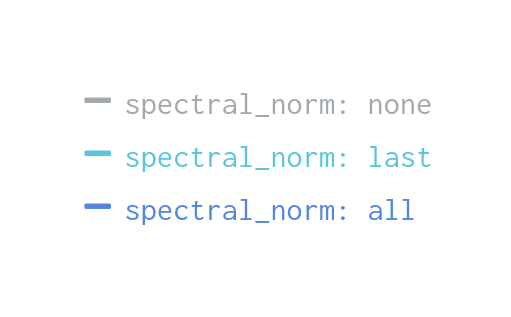}
    	\end{tabular}
    	\vspace*{7pt}
    	\caption{Comparison of the three variants of spectral normalization. Each curve shows the median over three seeds of the 100-episode running average of episode returns as a function of environment interactions.}
    	\label{tab:SN}
    \end{figure}
    
    Furthermore, we experimented with a fourth variant that applied SN to the final (noisy) linear layers only. However, we quickly dismissed this variant due to bad overall performance. We hypothesize that the drop in performance resulted from interference between spectral normalization and Noisy-Nets DQN exploration.
	
	Figure \ref{tab:SN} shows the comparison of the three variants of spectral normalization over five games. Even though the \emph{last} variant was marginally faster in terms of training throughput, we eventually settled on the \emph{all} variant as it slightly outperformed the former.
    
	Overall, we observed that spectral normalization has the largest effect at the beginning of training across a number of games. It significantly reduces the time until initial learning progress is made. This effect was most strongly observable in games such as \emph{Breakout}, \emph{Seaquest} and \emph{Tennis} where initial progress can be particularly slow.
	
	\subsection{Improving Hardware Utilization}
	
	We additionally employ several practical modifications to the training process that aim to decrease the wall-clock training time --- and thus the required computational budget --- by maximizing hardware utilization and training throughput:
	\begin{itemize}
        \item Similarly to \citet{procgen}, we increase the batch size from 32 to 256. As suggested in \citet{acceleratedrl}, we accordingly adjust the $\epsilon$ hyper-parameter for the Adam optimizer to $0.005/b$, where $b$ is the batch size.
        \item Environment interactions need to be computed sequentially and can thus not be parallelized across the time dimension. Thus, to batch environment simulation, we maintain $e = 64$ instances of the environment and take one step of this vectorized environment for every $k = 2$ training steps. In our implementation, the parameters $b$ and $e$ can be freely chosen so as to maximize GPU and CPU utilization and the agent's performance. The average number of times each environment transition is sampled from the replay buffer $\frac{b k}{e} = 8$ remains fixed by setting $k$ accordingly.
        \item We perform environment simulation steps and training steps in parallel to minimize idle time on the GPU.
        \item We use mixed-precision training as provided by PyTorch's \texttt{amp} package.
    \end{itemize}
    
\begin{figure}[h]
	\begin{center}
		\includegraphics[width=0.87\columnwidth]{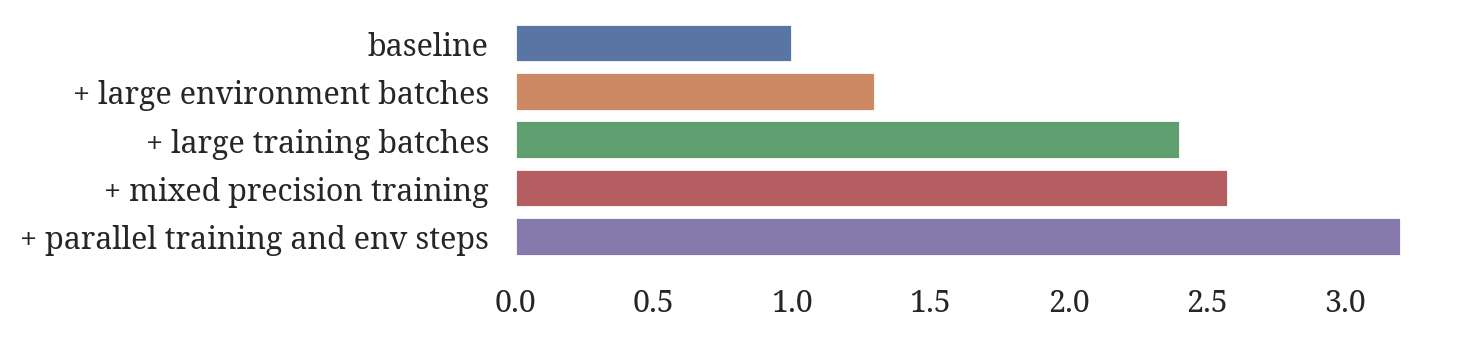}
	\end{center}\caption{Cumulative improvement in training throughput compared to the baseline.}
	\label{fig:speed}
\end{figure}

	In aggregate, these modifications decrease the training time by a factor of 3.2, from 24 to approximately 7.5 hours, for training on 10M frames on a single Nvidia RTX 3090 GPU. Interestingly, the increased batch size also considerably improves the performance on several games. Prior work has observed a similar effect and attributed it to decreased gradient variance and thus more stable training behavior \citep{procgen}. Figure \ref{fig:speed} shows the individual contributions of the modifications compared to baseline.
	
    \subsection{Hyperparameter Tuning}
	Apart from the batch size, network size, and choice of spectral normalized layers, we also performed a hyper-parameter search over learning rates and compared the Huber and mean-squared error (MSE) loss functions.
	
	Unlike \citet{revrainbow} we did not observe better performance through the use of the MSE loss. However, the search over learning rates did prove fruitful. Increasing the learning rate by a factor of 4 (to the same value used in DQN) significantly speeds up learning in our experiments, both with the increased batch size and without. In addition, it considerably improves the final performance the agent achieves. We hypothesize that the reason \citet{rainbow} settled on the lower learning rate was that the benefit of better convergence when using a lower learning rate outweighed the disadvantage of slower learning when training for much longer periods of time.
	
	The full set of hyperparameters is listed in Appendix A in the supplementary material.
	
	\subsection{Distributional RL}
	Finally, in our experiments removing the distributional RL component from Rainbow did not significantly affect our agent's final performance or learning speed when trained for 10M frames. This is in line with observations from the ablation studies in \citet{rainbow} that showed the main benefit of distributional RL materialized only after training for about 40M frames. Thus, considering the negligible performance improvement and substantial implementation complexity, we decided not to include distributional RL in our final agent.
	
	Similarly, the evaluation curves in \citet{rainbow} indicate that other components of Rainbow such as Noisy-Nets DQN and Dueling DQN may only be important when training for long periods. This may suggest that --- given future improvements in data efficiency, leading to lower required training time --- it might be possible to further simplify the Rainbow agent with no penalties to performance, in particular when working with a limited computational budget.
	
	Additionally, we investigated replacing the C51 \citep{distributional} variant of distributional RL with QR-DQN \citep{qrdqn}. However, in our experiments, we could not observe a significant difference. 
	
	\section{Results}
	Table \ref{tab:overallres} compares the final aggregate scores of our approach to the ones of the original Rainbow-DQN agent, as well as to standard DQN. The full scores and learning curves for each of the 53 games are available in Table 1 and Figure 2 in the supplementary material.
	
    Each of our training runs took approximately 7.5 hours on a single Nvidia RTX 3090 GPU or 10 hours on a single Nvidia 2080Ti GPU. These requirements make performing larger numbers of experiments feasible, even with a more modest compute budget. 
	
	Our final agent was trained for only 10 million environment transitions and achieves a median human normalized score (HNS) of 205.7, reaching above human-level performance on 39 out of 53 games. In contrast, classic Rainbow achieves a median HNS of 70 (estimated) and 231 after training for 10 and 200 million transitions, respectively. Overall, their score after 200M transitions is equivalent to super-human performance on 40 out of 53 games. 
	The exact scores for classic Rainbow at 10M frames were not published in \citet{rainbow}. Therefore, they were estimated from the provided evaluation curves. In addition, published results in \citet{dopamine} for the "Dopamine" Rainbow implementation serve as another reference point (see Table \ref{tab:overallres}, HNS of 50.4 for 10M and 145.8 for 200M). In any case, our approach outperforms the scores at 10M by a big margin and achieves scores more similar to the ones obtained after 200M steps.  

\begin{table}[!htp]\centering
    \renewcommand{\arraystretch}{1.2}
    \scriptsize
    \begin{tabular}{l >{\raggedleft}m{7.5em} >{\raggedleft}m{7.5em} >{\raggedleft}m{7.5em} >{\raggedleft}m{7.5em} >{\raggedleft\arraybackslash}m{7.5em}}
 &\textbf{Rainbow \citep{dopamine}} &\textbf{DQN \citep{mnih2013}} &\textbf{Rainbow \citep{dopamine}} &\textbf{Rainbow \citep{rainbow}} &\textbf{Ours} \\
 
\textbf{training frames} &\cellcolor[HTML]{c9daf8}10M &\cellcolor[HTML]{ead1dc}200M &\cellcolor[HTML]{ead1dc}200M &\cellcolor[HTML]{ead1dc}200M &\cellcolor[HTML]{c9daf8}10M \\
\textbf{mean HNS} &\cellcolor[HTML]{fff2cc}166.9 &\cellcolor[HTML]{f9f0ca}224.2 &\cellcolor[HTML]{8dcd9f}1161.6 &\cellcolor[HTML]{57bb8a}1624.9 &\cellcolor[HTML]{8bcd9f}1174.2 \\
\textbf{median HNS} &\cellcolor[HTML]{fff2cc}50.4 &\cellcolor[HTML]{e5eac2}79.3 &\cellcolor[HTML]{a7d5aa}145.8 &\cellcolor[HTML]{57bb8a}231.0 &\cellcolor[HTML]{6fc394}205.7 \\
\textbf{\# games above human} &\cellcolor[HTML]{fff2cc}19 &\cellcolor[HTML]{d7e5bd}24 &\cellcolor[HTML]{67c191}38 &\cellcolor[HTML]{57bb8a}40 &\cellcolor[HTML]{5fbe8e}39 \\
    \end{tabular}
    \vspace*{10pt}
    \caption{Comparison of human normalized scores averaged over the 53 tested games (and 3 random seeds) for our agent, two implementations of Rainbow, and DQN. Random and human scores used for computing the HNS were taken from \citet{mnih2015} where available, otherwise from \citet{agent57}. The full results are provided in the supplementary material.}
    \label{tab:overallres}
\end{table}
	
	\section{Conclusion}
	
	In this paper, we addressed one of the fundamental questions in reinforcement learning: how to accelerate the training of RL agents? To achieve this, we employed a variety of tricks that speed up the RL training procedure and empirically investigated how they affect training speed as well as overall performance on the Atari benchmark. While most of these enhancements have been studied individually in the context of RL, our work is, to the best of our knowledge, the first to study them jointly. Our improved training procedure significantly reduces the training time and data required by the RL agent while maintaining competitive performance. Furthermore, we conducted a number of ablations to understand what effect the individual modifications have on the training process. Overall, the results we presented may serve as a practical guide for speeding up the training of RL agents.

    \begin{ack}
    	We would like to thank the TU Wien DataLab for providing the majority of the needed compute resources and the CleanRL team for their baseline implementation of DQN \citep{cleanrl}.
    \end{ack}
    
    \newpage
    
	\bibliographystyle{plainnat}
	\bibliography{biblio}
    \appendix
    
    \appendix
    
\newpage
\section{Full Results}\label{sec:results}
	\begin{figure}[!h]
    	\renewcommand{\arraystretch}{4}
    	\setlength\tabcolsep{0pt}
        \centering
    	\begin{tabular}{lllll}
    	\includegraphics[width=.18\linewidth,valign=m]{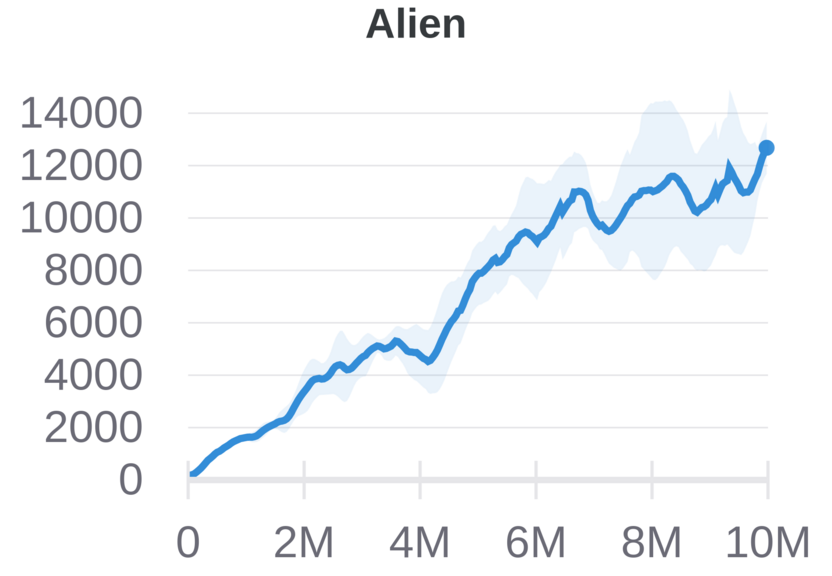} &
    	\includegraphics[width=.18\linewidth,valign=m]{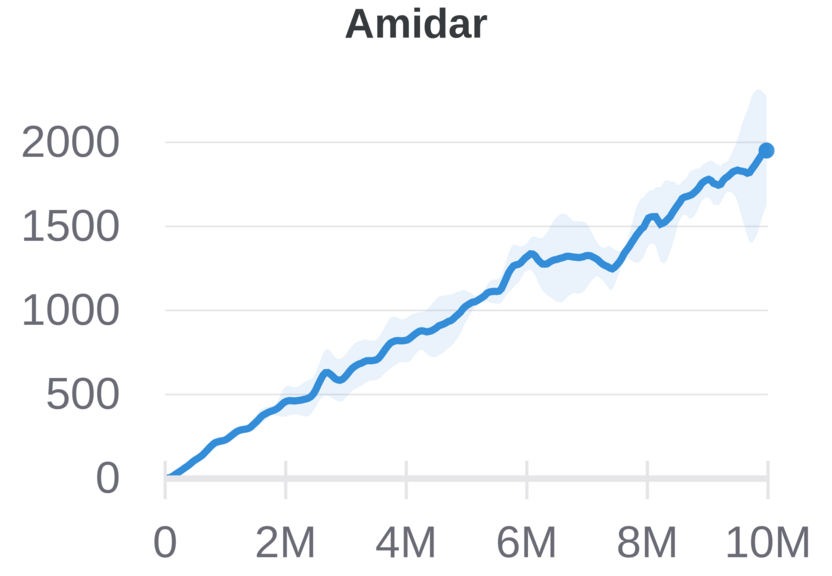} & 
    	\includegraphics[width=.18\linewidth,valign=m]{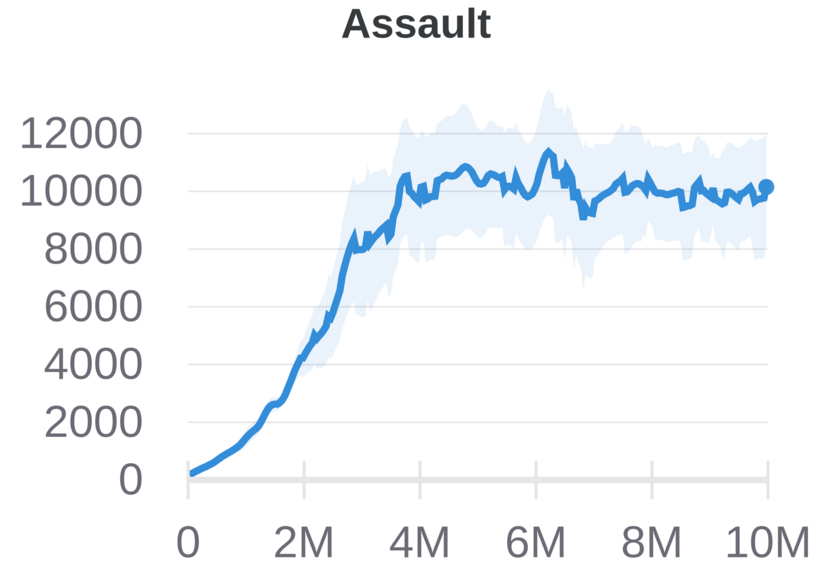} & 
    	\includegraphics[width=.18\linewidth,valign=m]{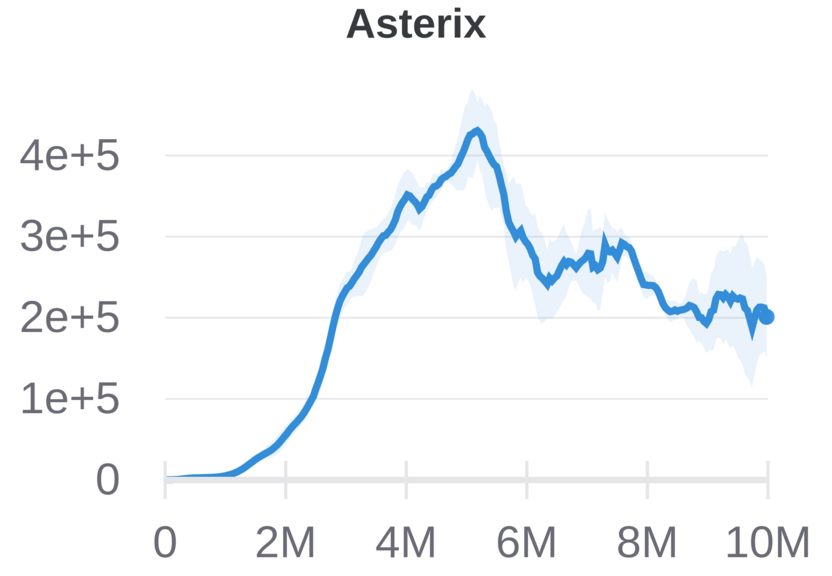} & 
    	\includegraphics[width=.18\linewidth,valign=m]{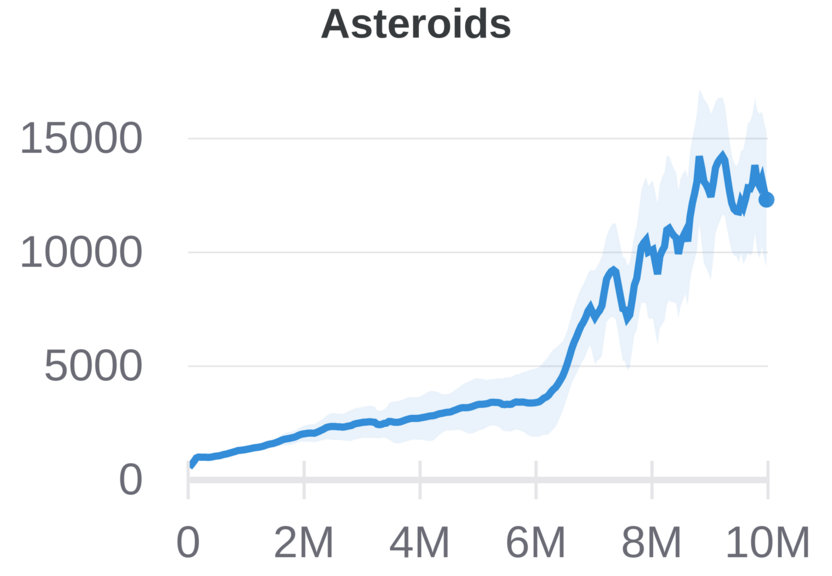} \\

    	\includegraphics[width=.18\linewidth,valign=m]{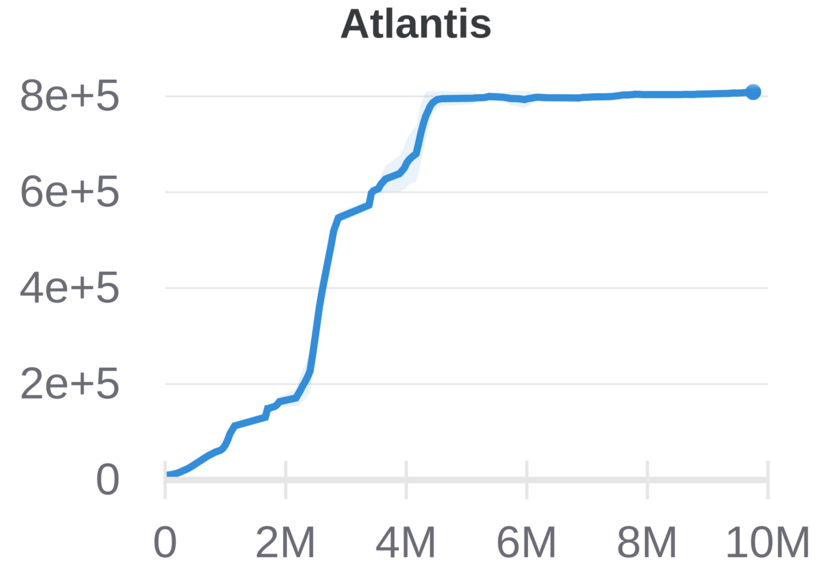} &
    	\includegraphics[width=.18\linewidth,valign=m]{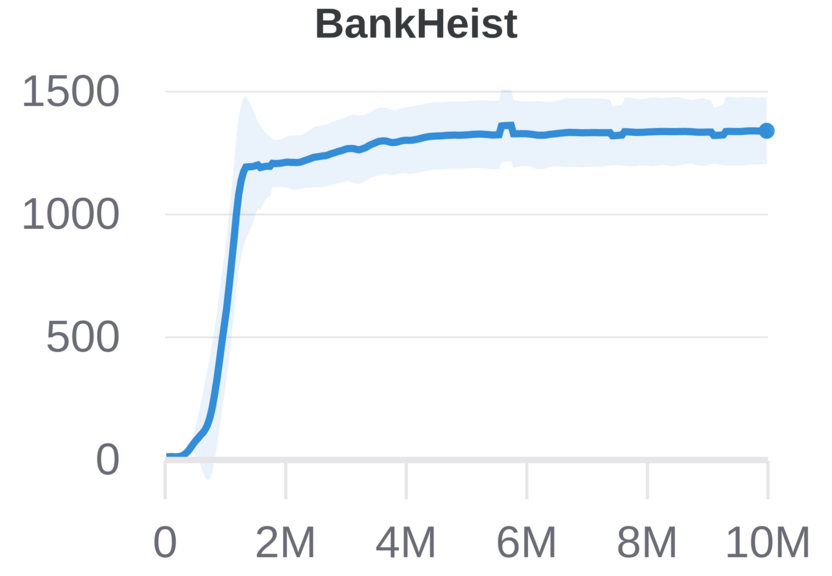} & 
    	\includegraphics[width=.18\linewidth,valign=m]{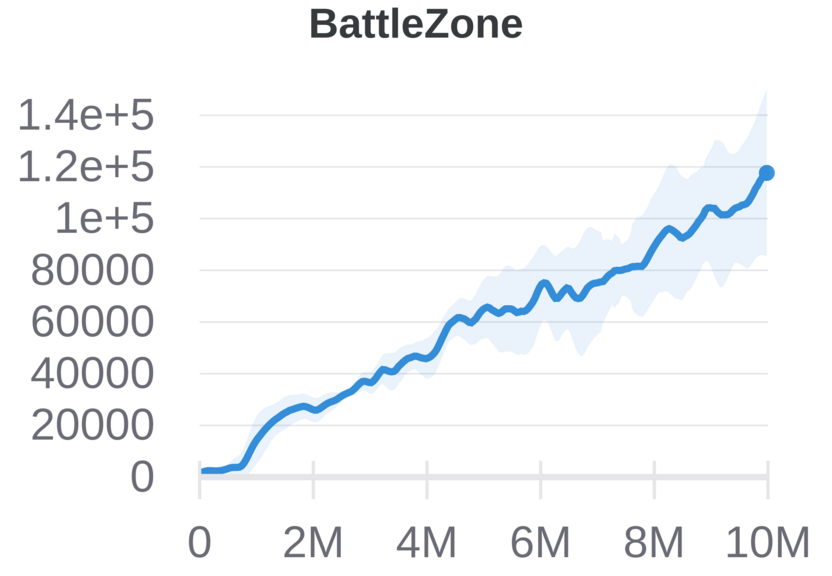} & 
    	\includegraphics[width=.18\linewidth,valign=m]{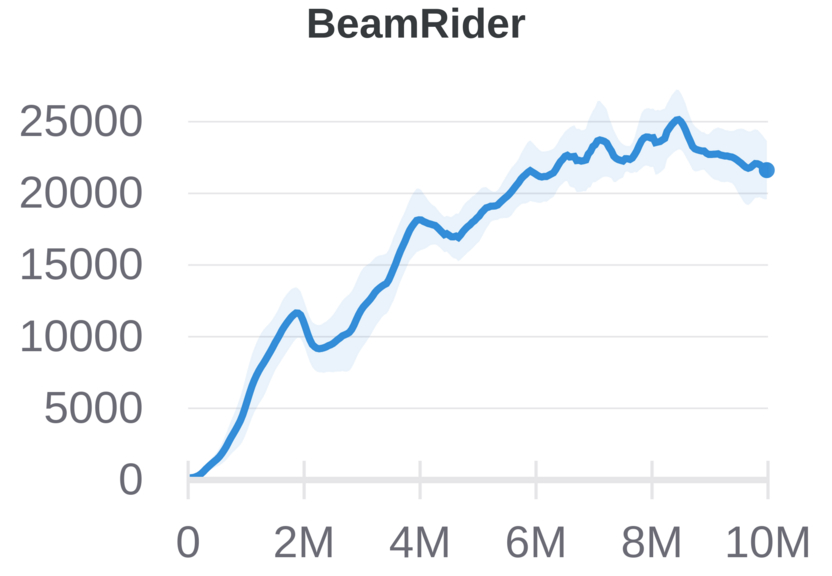} & 
    	\includegraphics[width=.18\linewidth,valign=m]{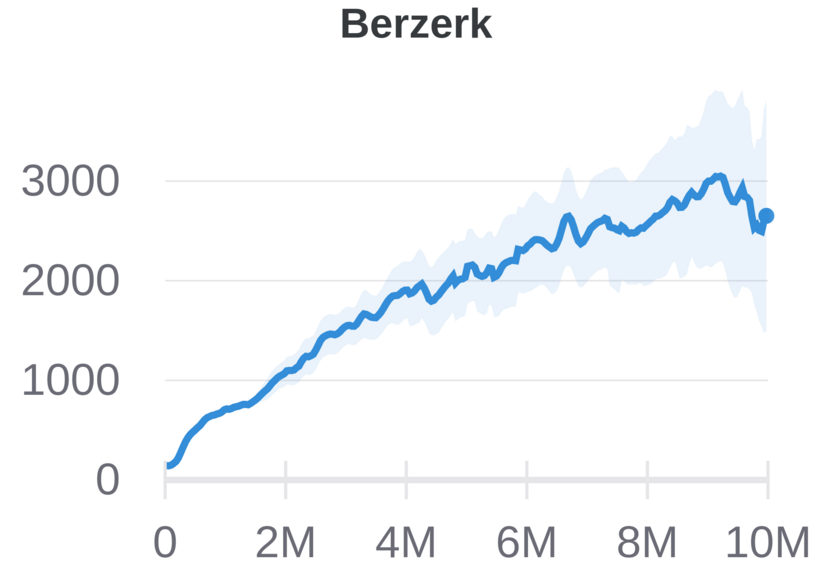} \\

    	\includegraphics[width=.18\linewidth,valign=m]{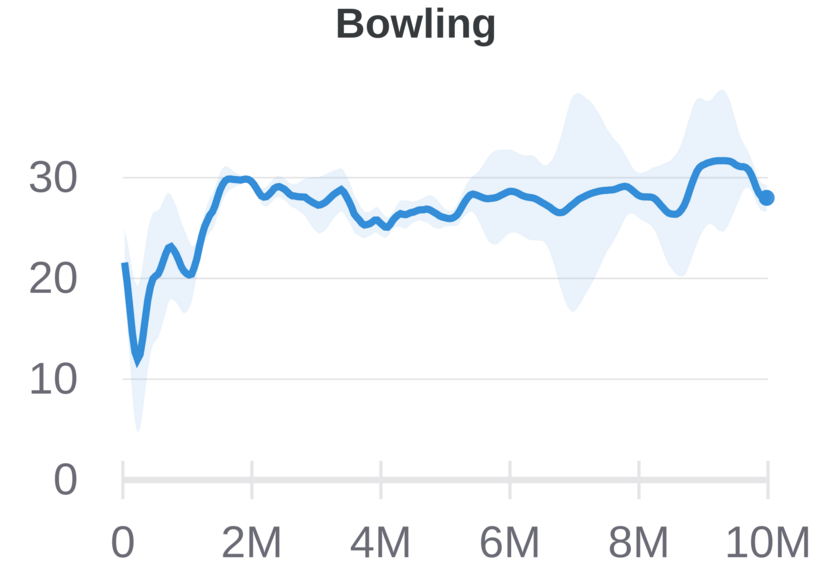} &
    	\includegraphics[width=.18\linewidth,valign=m]{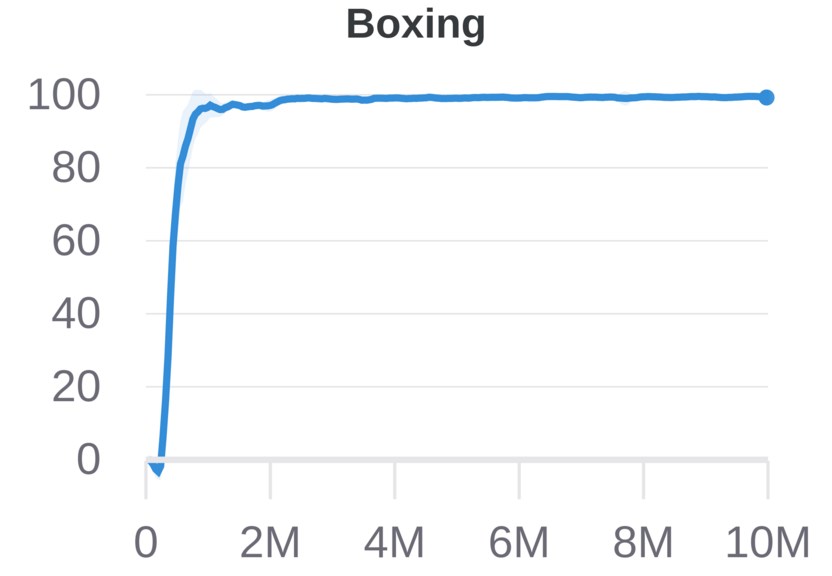} & 
    	\includegraphics[width=.18\linewidth,valign=m]{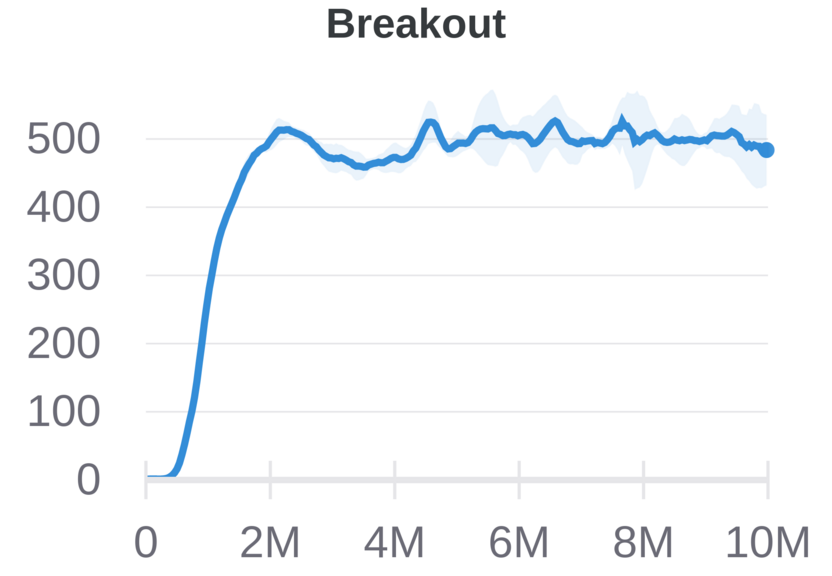} & 
    	\includegraphics[width=.18\linewidth,valign=m]{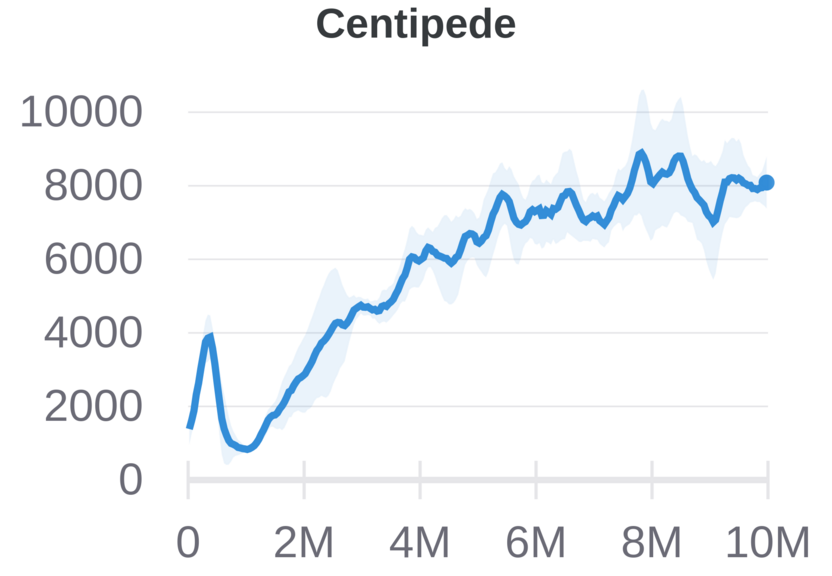} & 
    	\includegraphics[width=.18\linewidth,valign=m]{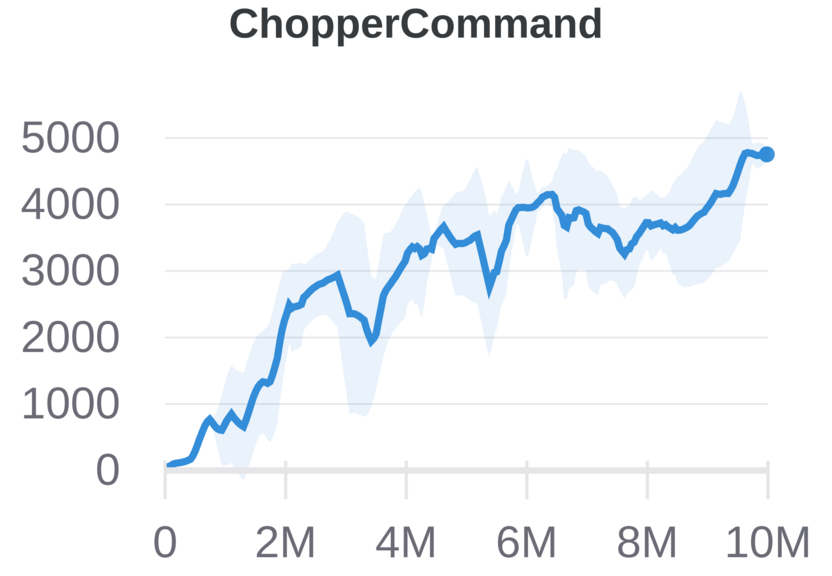} \\

    	\includegraphics[width=.18\linewidth,valign=m]{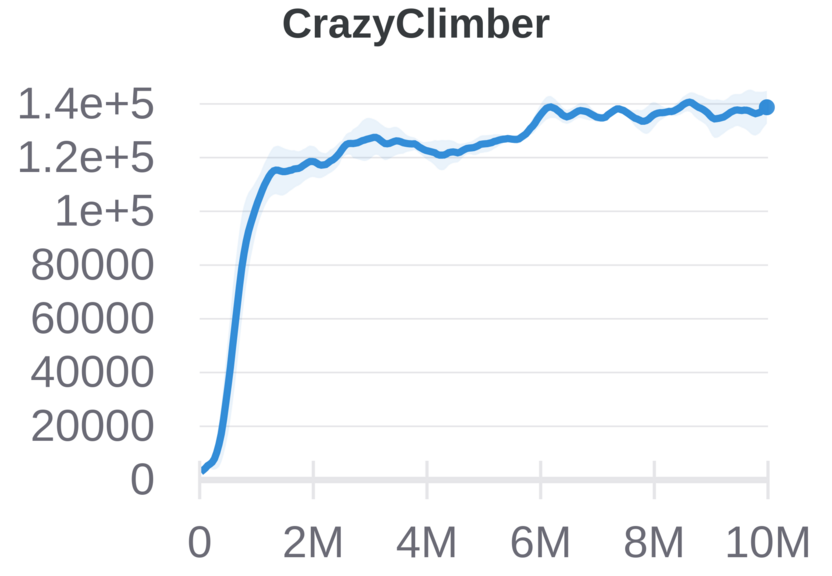} &
    	\includegraphics[width=.18\linewidth,valign=m]{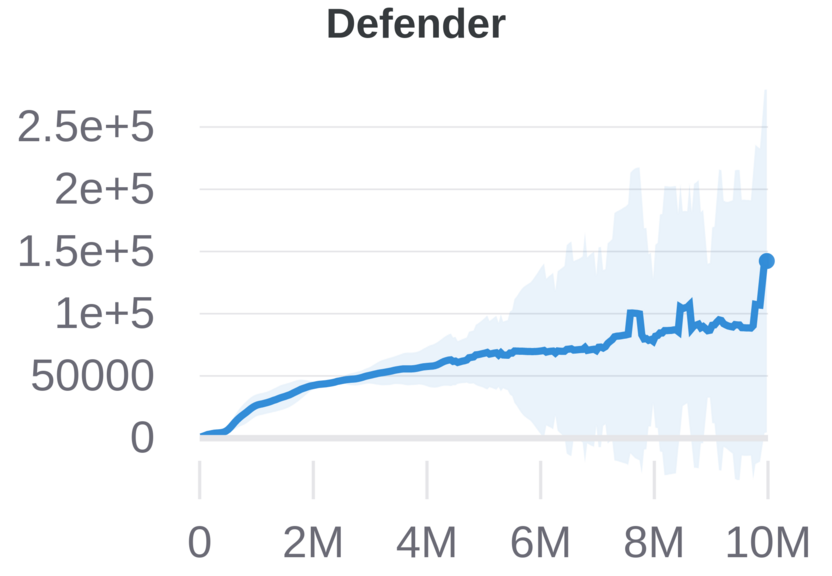} & 
    	\includegraphics[width=.18\linewidth,valign=m]{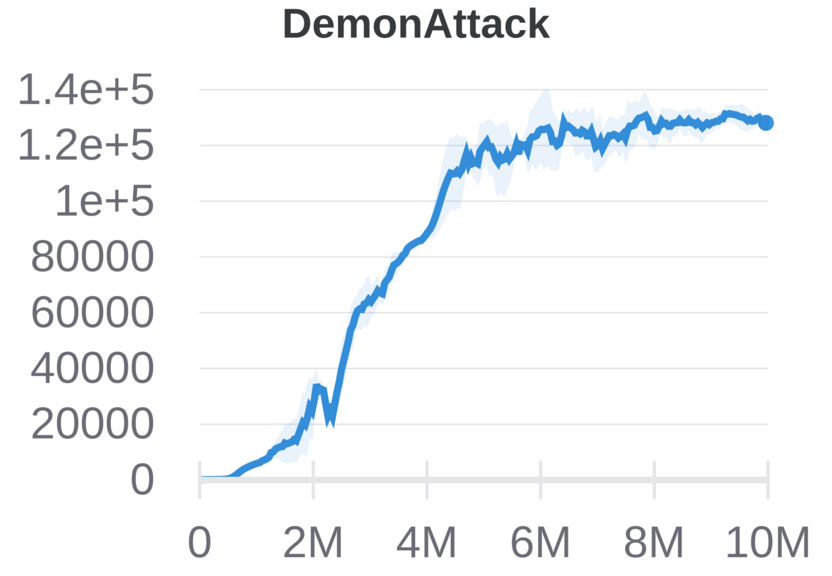} & 
    	\includegraphics[width=.18\linewidth,valign=m]{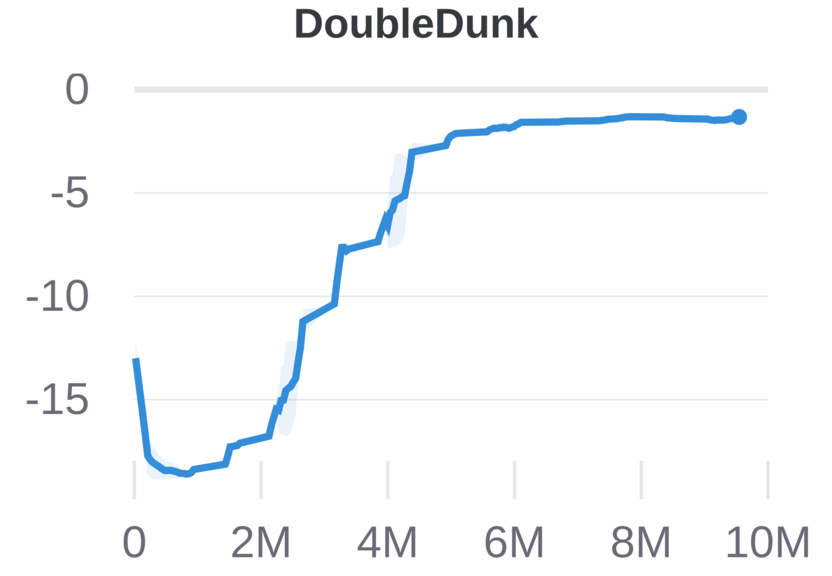} & 
    	\includegraphics[width=.18\linewidth,valign=m]{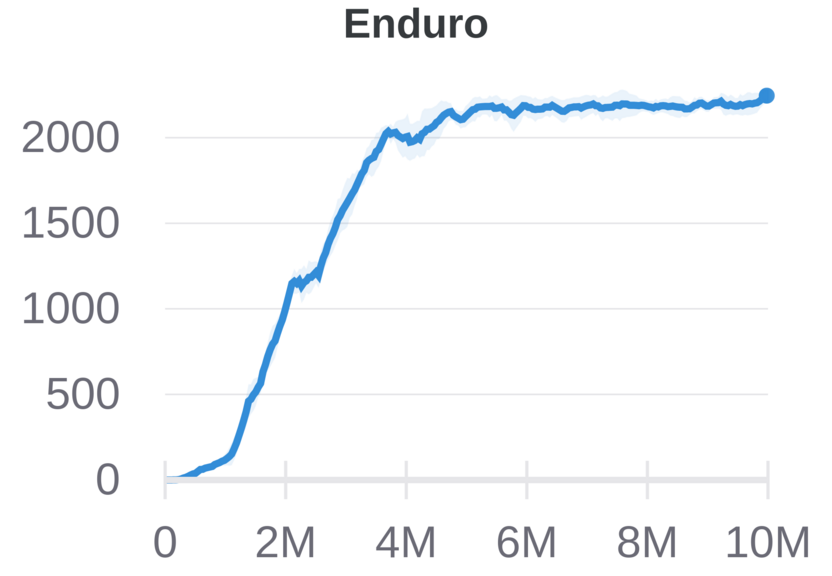} \\

    	\includegraphics[width=.18\linewidth,valign=m]{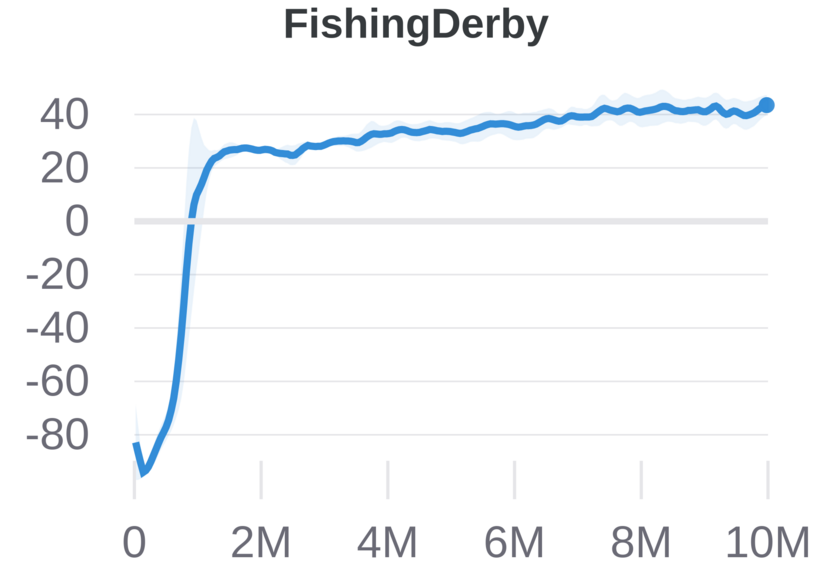} &
    	\includegraphics[width=.18\linewidth,valign=m]{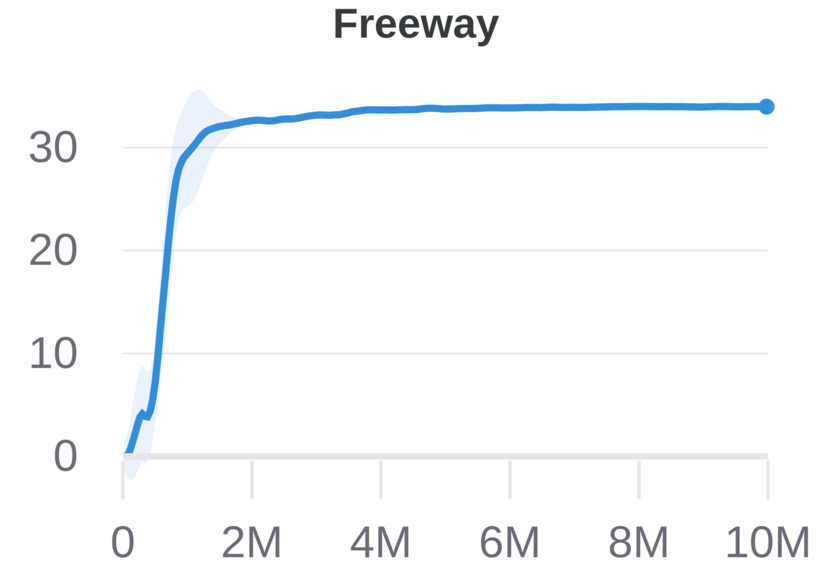} &
    	\includegraphics[width=.18\linewidth,valign=m]{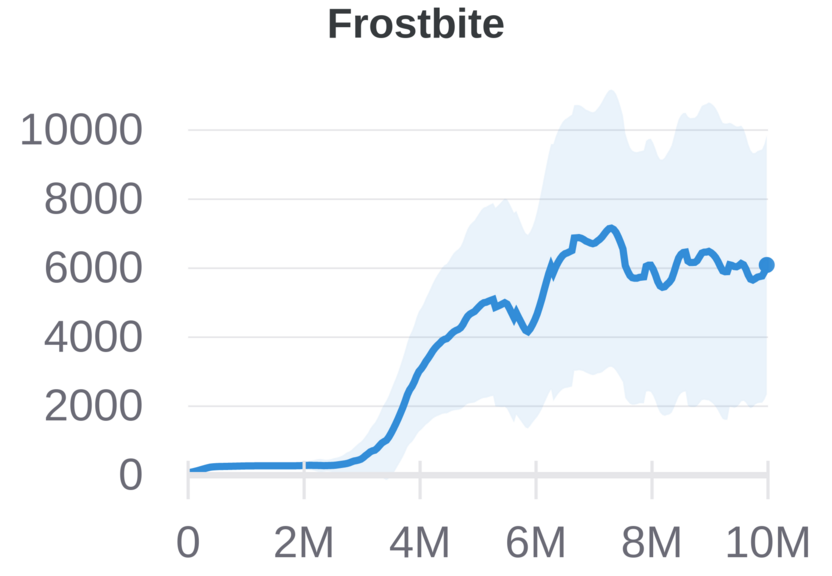} & 
    	\includegraphics[width=.18\linewidth,valign=m]{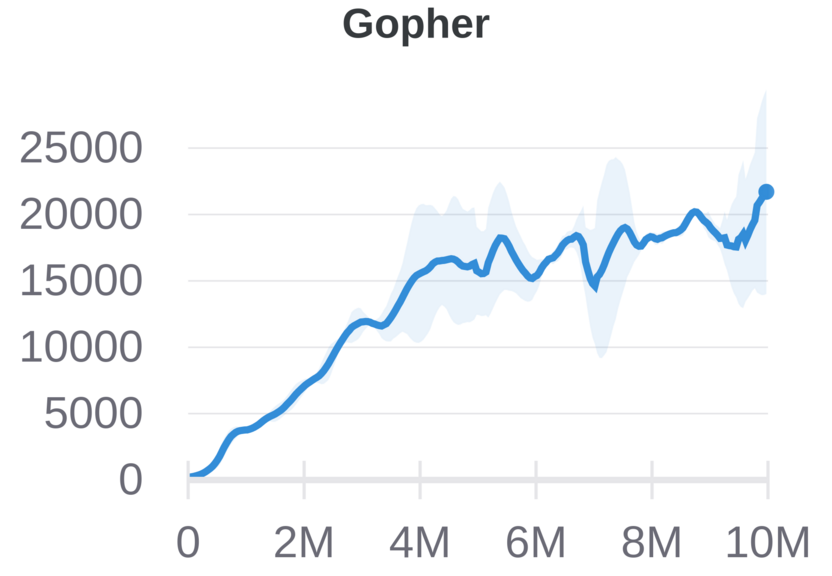} &
    	\includegraphics[width=.18\linewidth,valign=m]{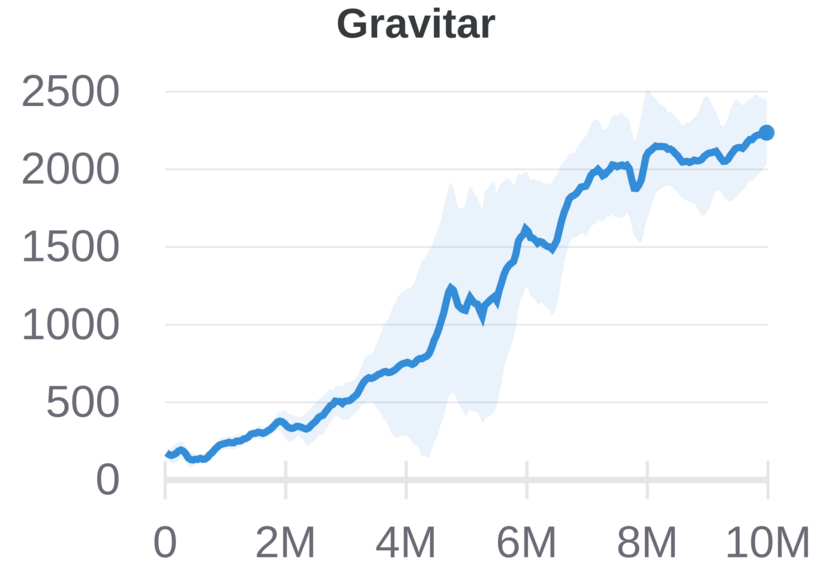} \\

    	\includegraphics[width=.18\linewidth,valign=m]{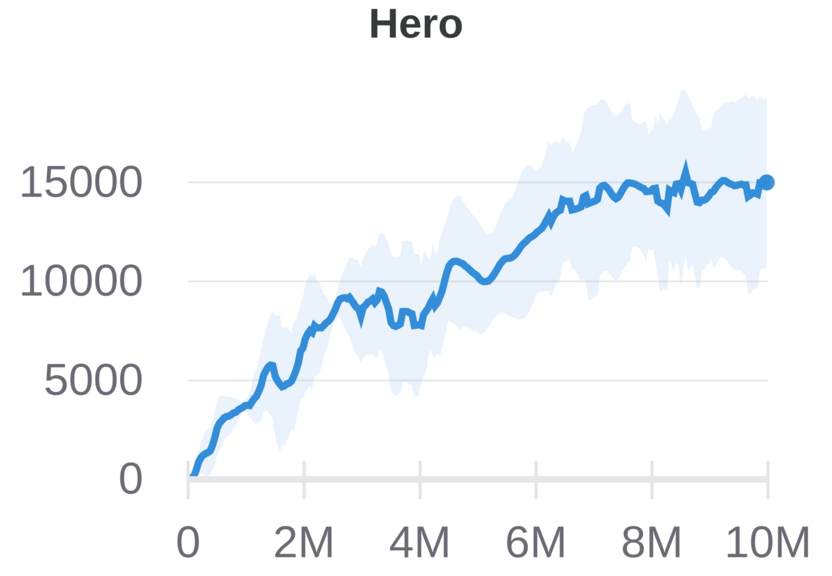} & 
    	\includegraphics[width=.18\linewidth,valign=m]{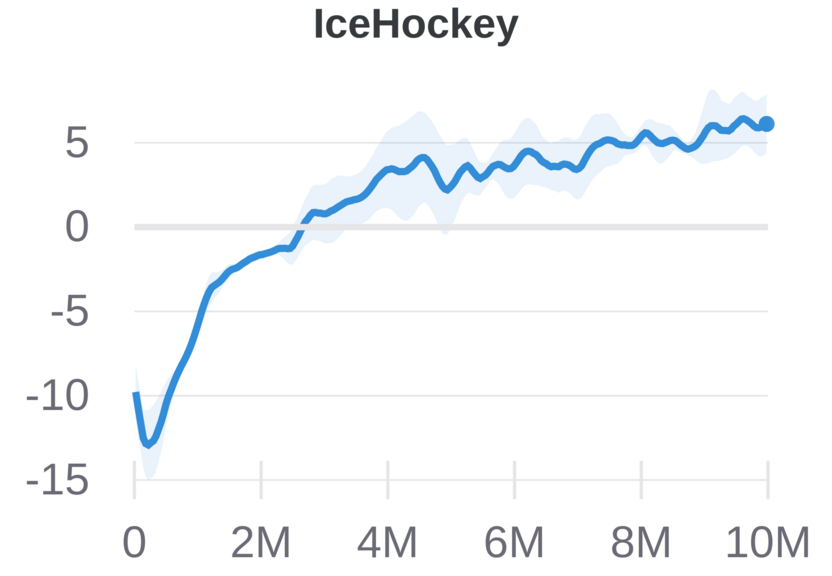} &
    	\includegraphics[width=.18\linewidth,valign=m]{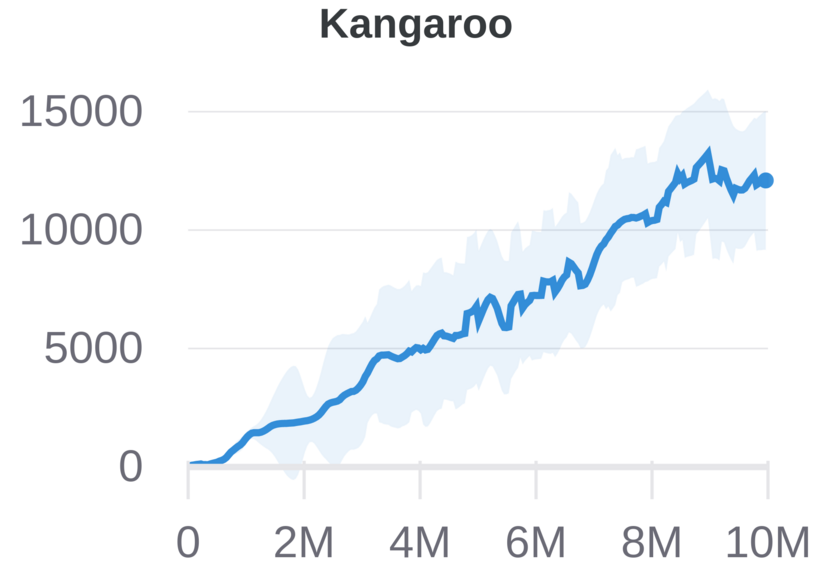} &
    	\includegraphics[width=.18\linewidth,valign=m]{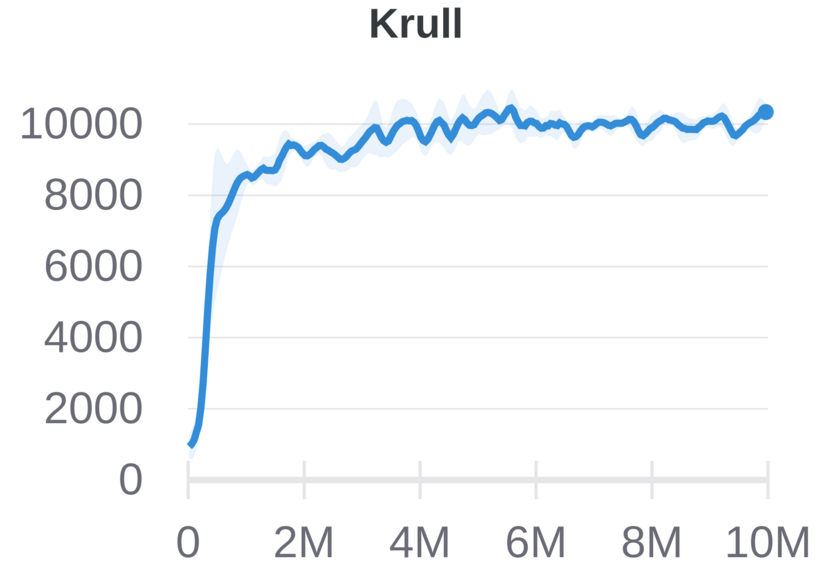} & 
    	\includegraphics[width=.18\linewidth,valign=m]{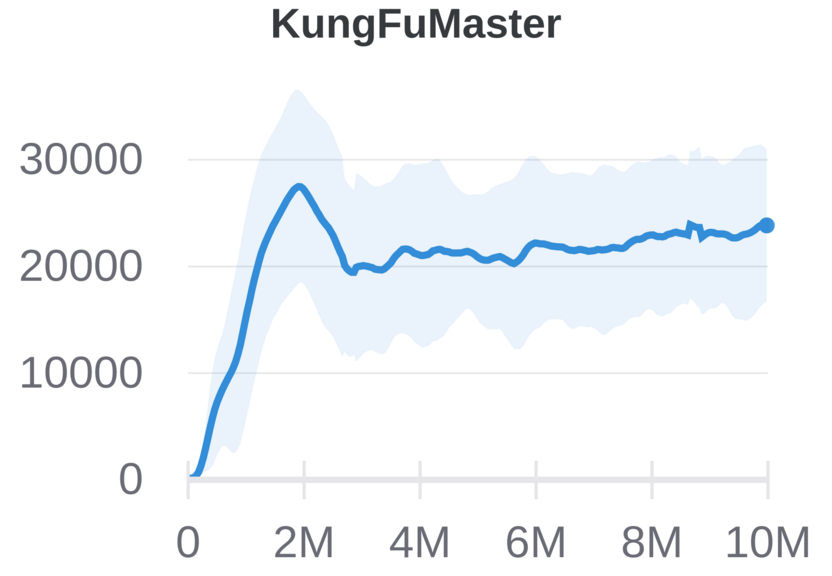} \\

    	\includegraphics[width=.18\linewidth,valign=m]{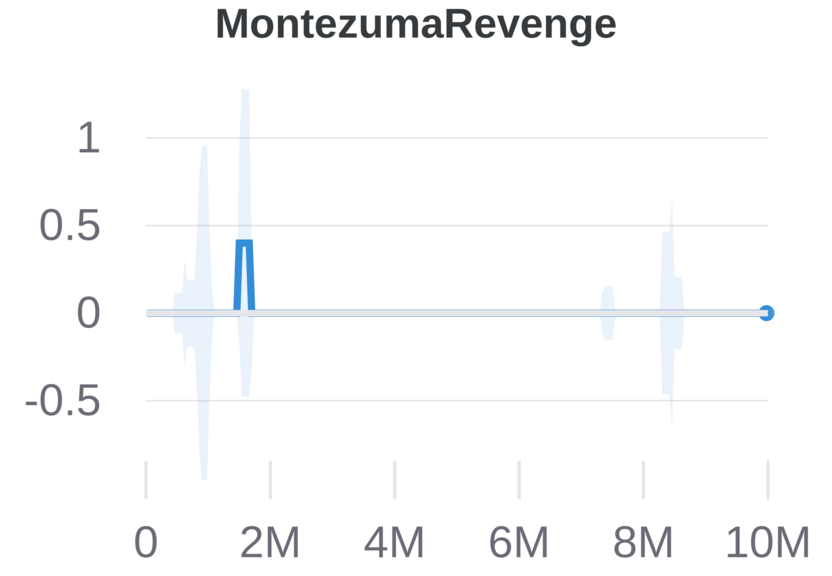} &
    	\includegraphics[width=.18\linewidth,valign=m]{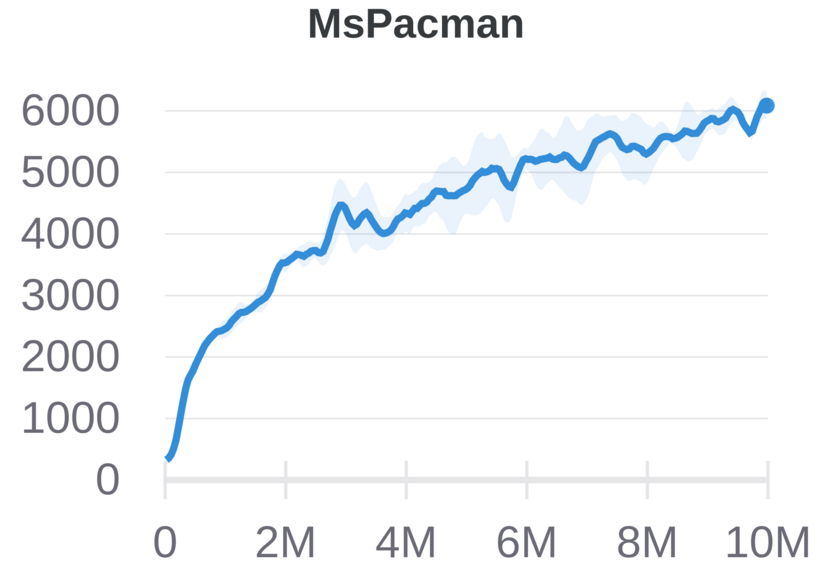} & 
    	\includegraphics[width=.18\linewidth,valign=m]{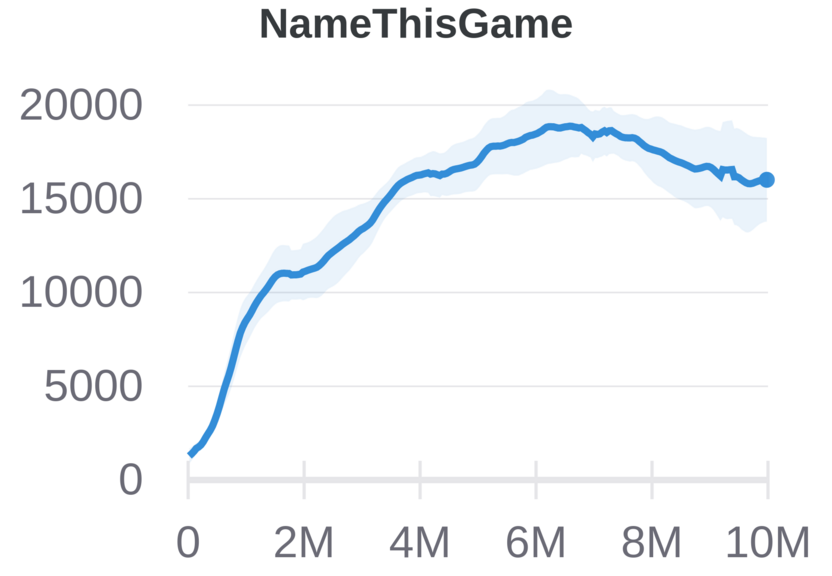} &
    	\includegraphics[width=.18\linewidth,valign=m]{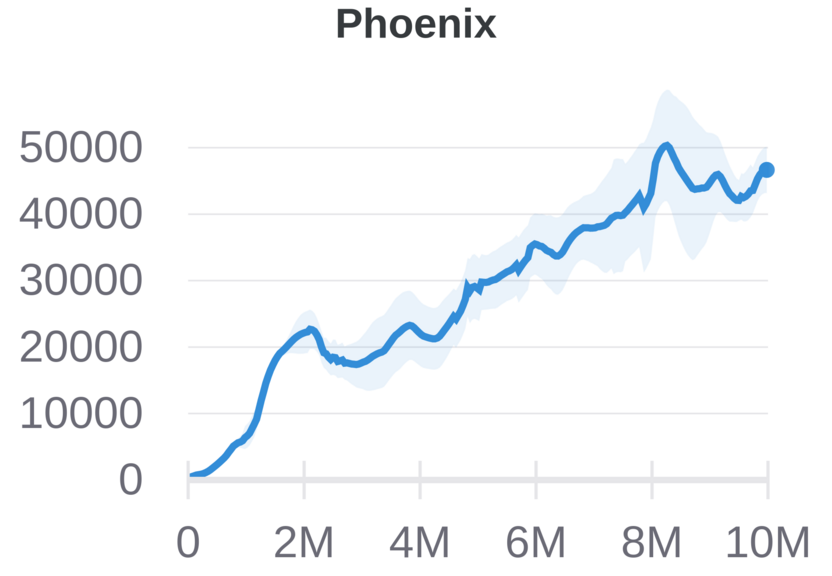} &
    	\includegraphics[width=.18\linewidth,valign=m]{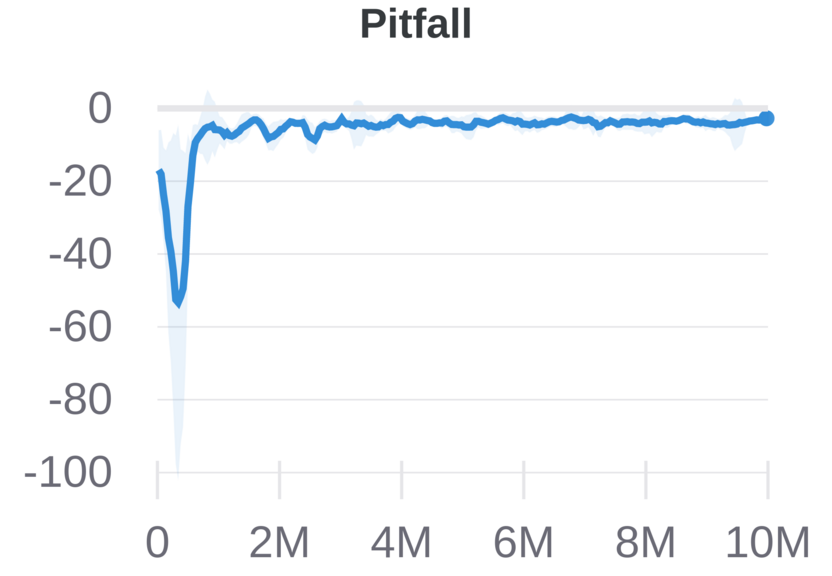} \\

    	\includegraphics[width=.18\linewidth,valign=m]{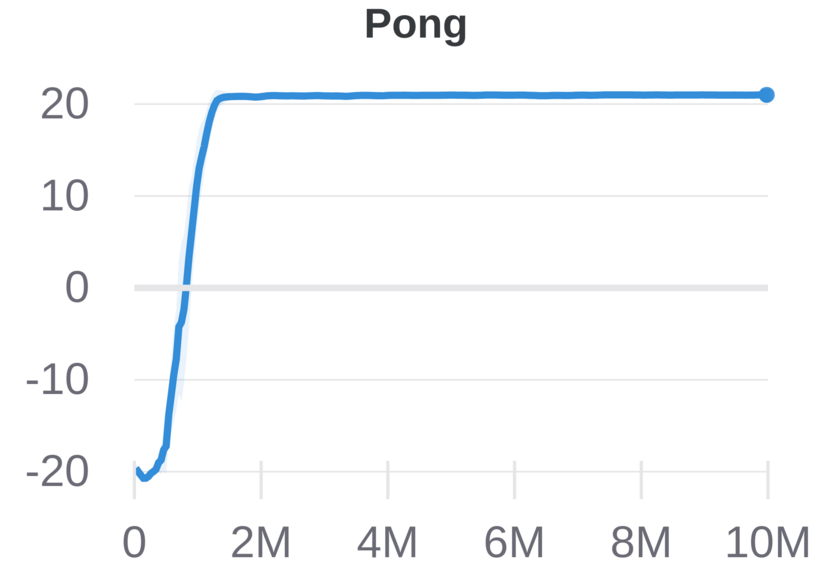} & 
    	\includegraphics[width=.18\linewidth,valign=m]{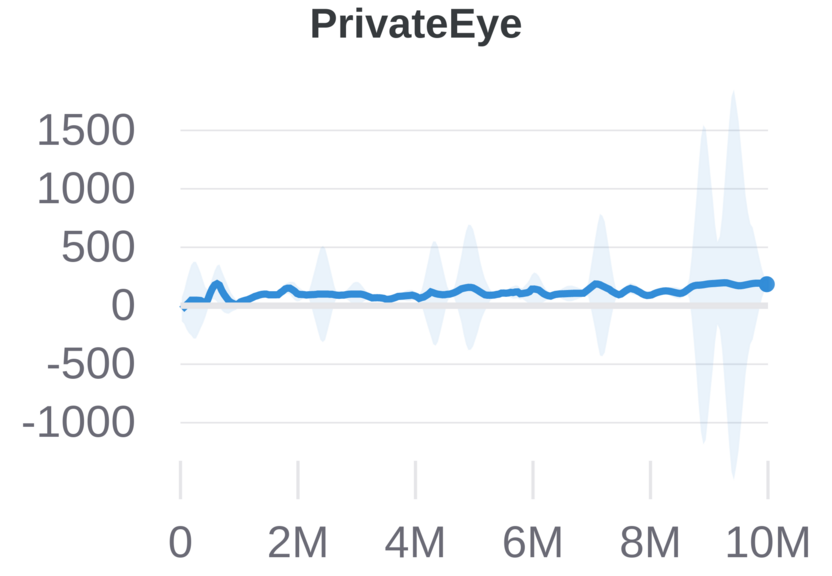} & 
    	\includegraphics[width=.18\linewidth,valign=m]{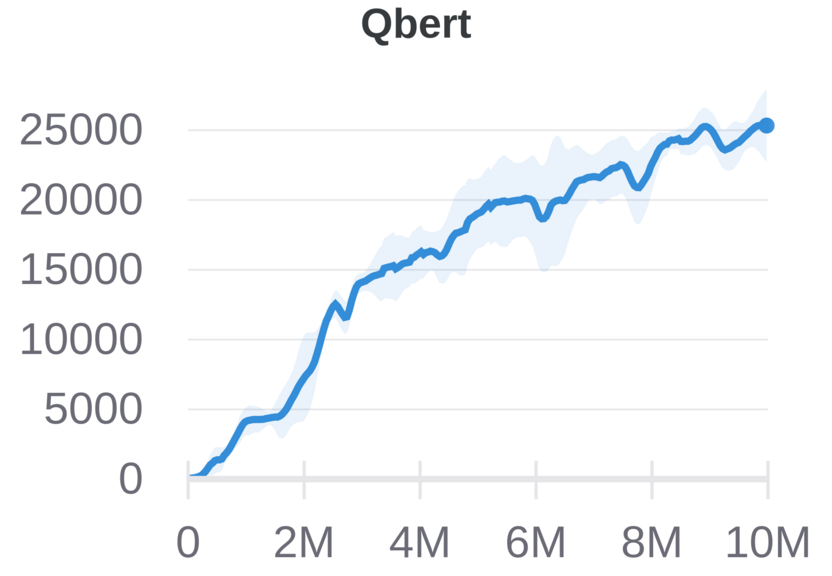} &
    	\includegraphics[width=.18\linewidth,valign=m]{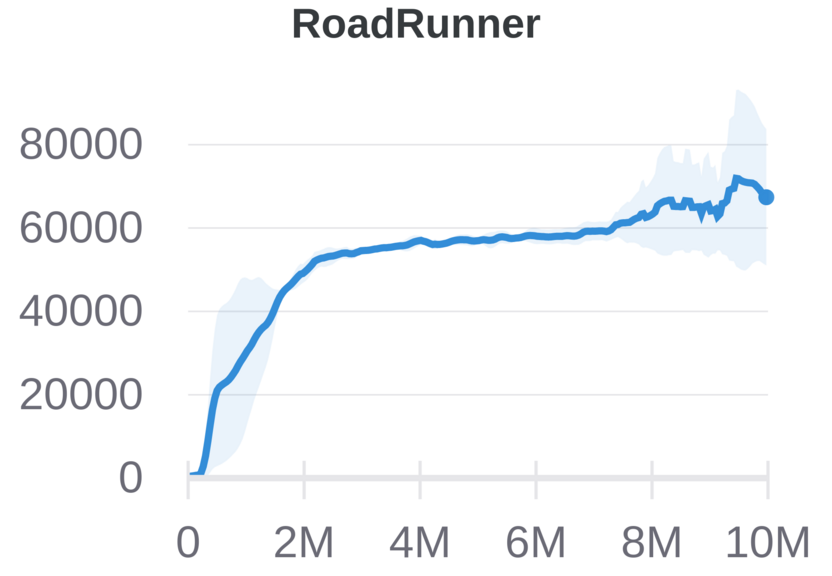} &
    	\includegraphics[width=.18\linewidth,valign=m]{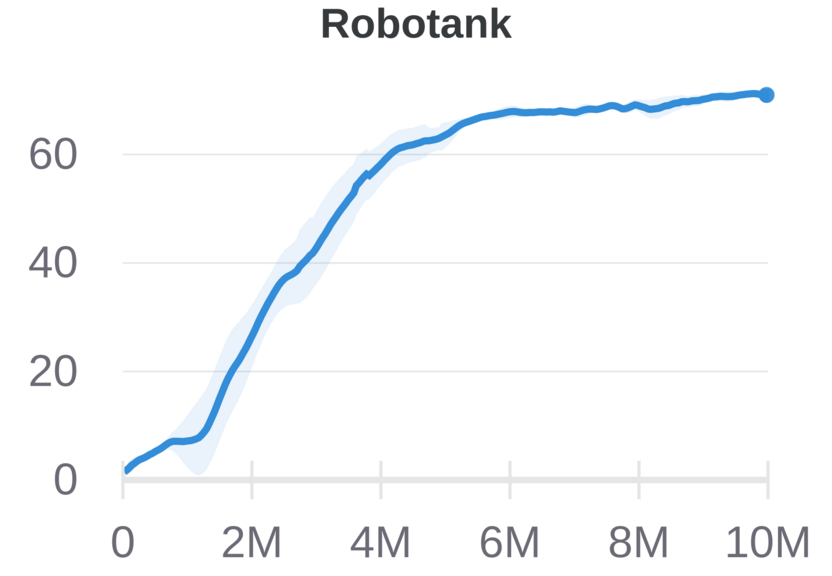} \\

    	\includegraphics[width=.18\linewidth,valign=m]{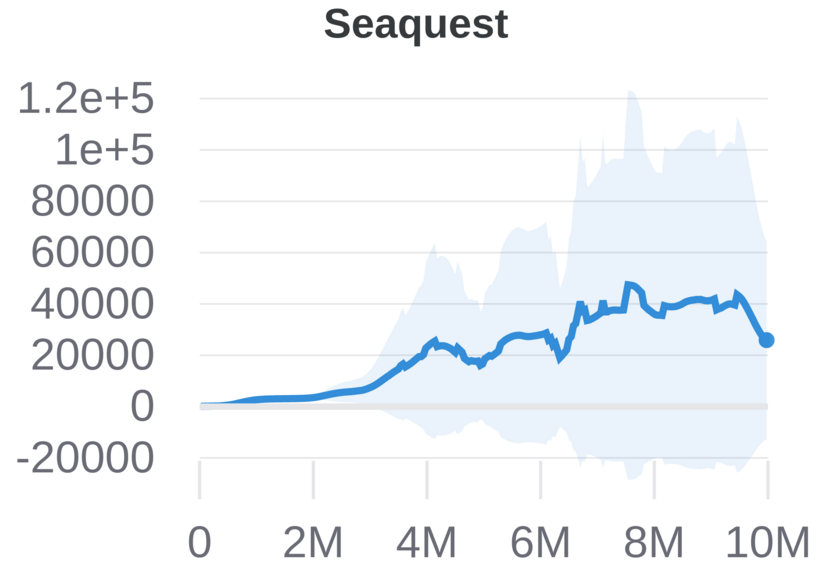} & 
    	\includegraphics[width=.18\linewidth,valign=m]{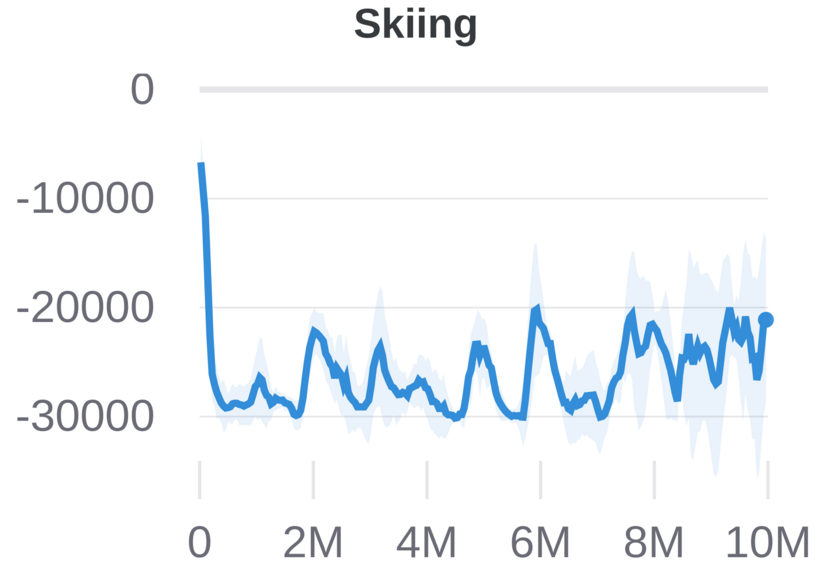} & 
    	\includegraphics[width=.18\linewidth,valign=m]{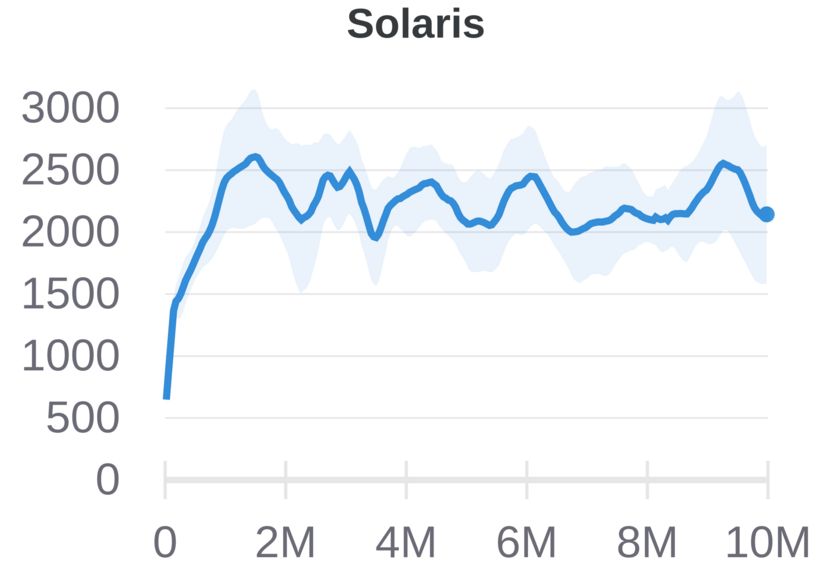} &
    	\includegraphics[width=.18\linewidth,valign=m]{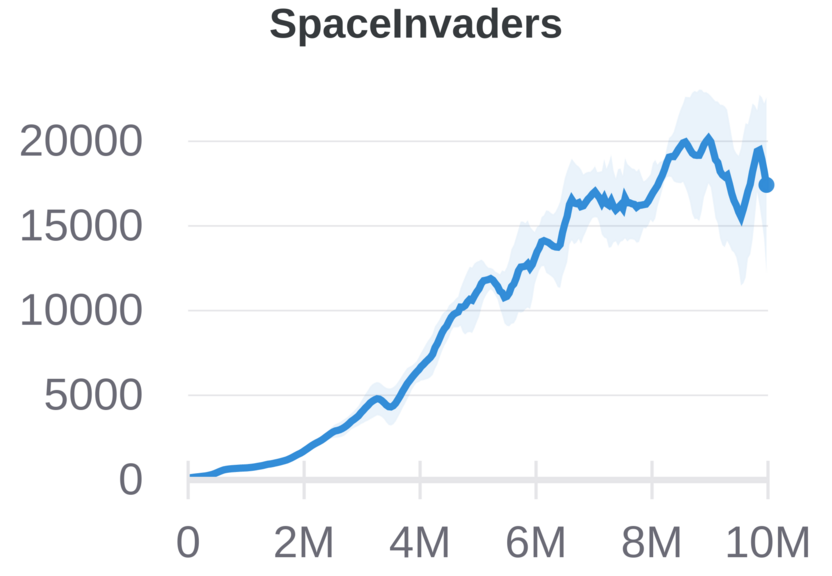} &
    	\includegraphics[width=.18\linewidth,valign=m]{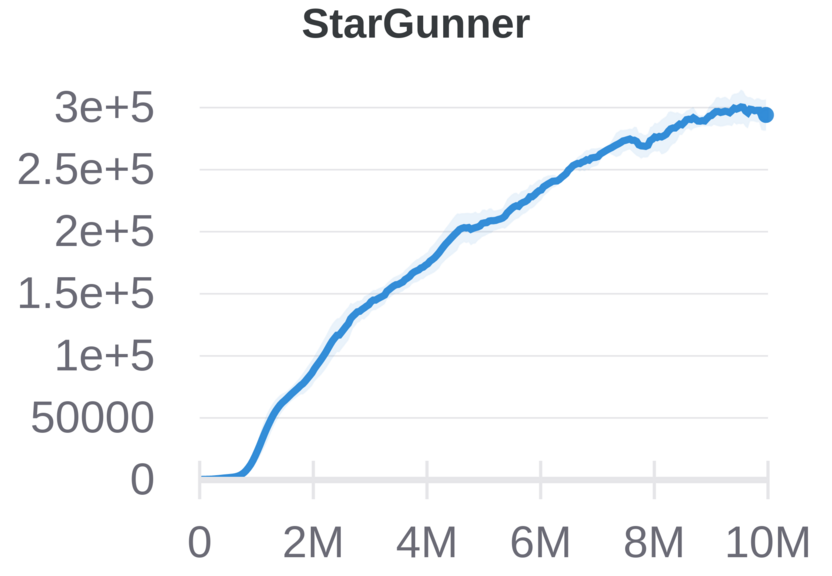} \\

    	\includegraphics[width=.18\linewidth,valign=m]{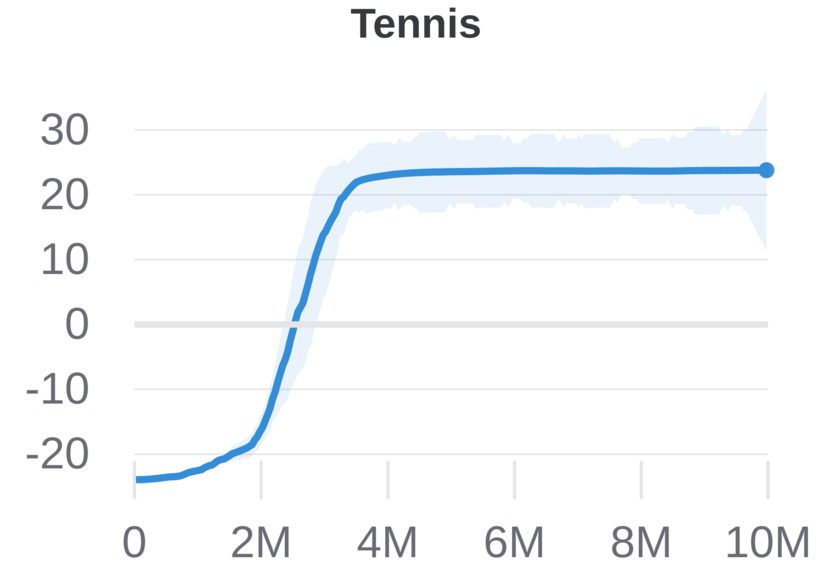} & 
    	\includegraphics[width=.18\linewidth,valign=m]{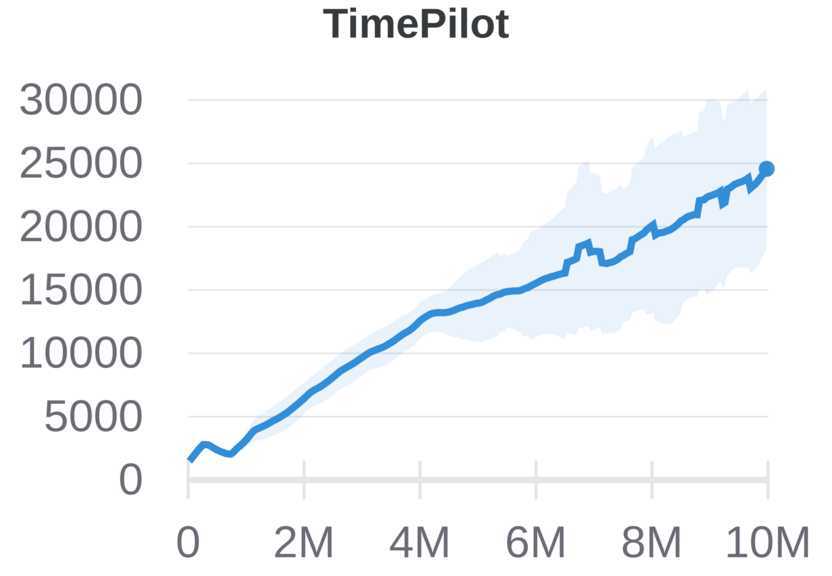} & 
    	\includegraphics[width=.18\linewidth,valign=m]{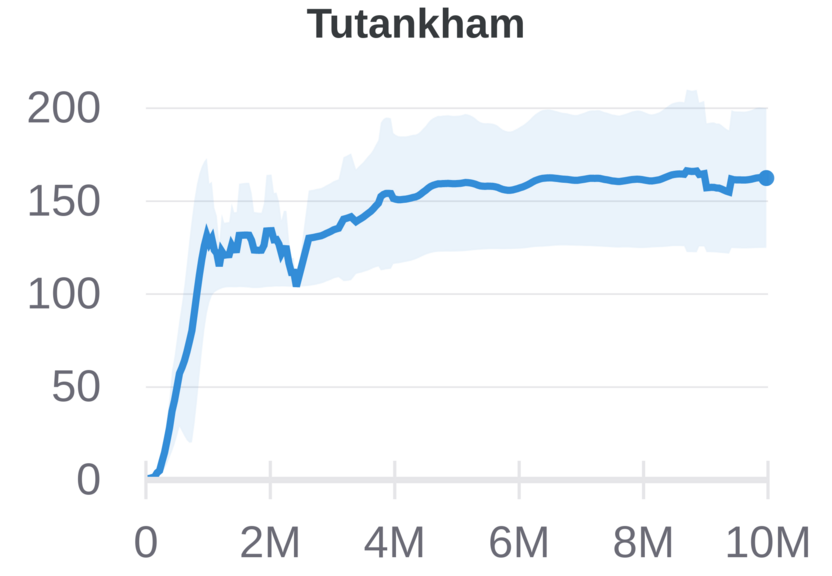} &
    	\includegraphics[width=.18\linewidth,valign=m]{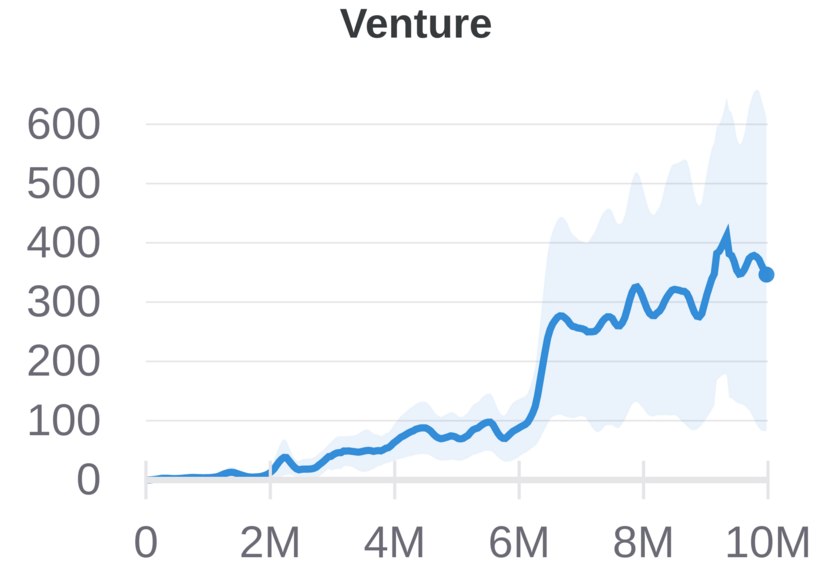} &
    	\includegraphics[width=.18\linewidth,valign=m]{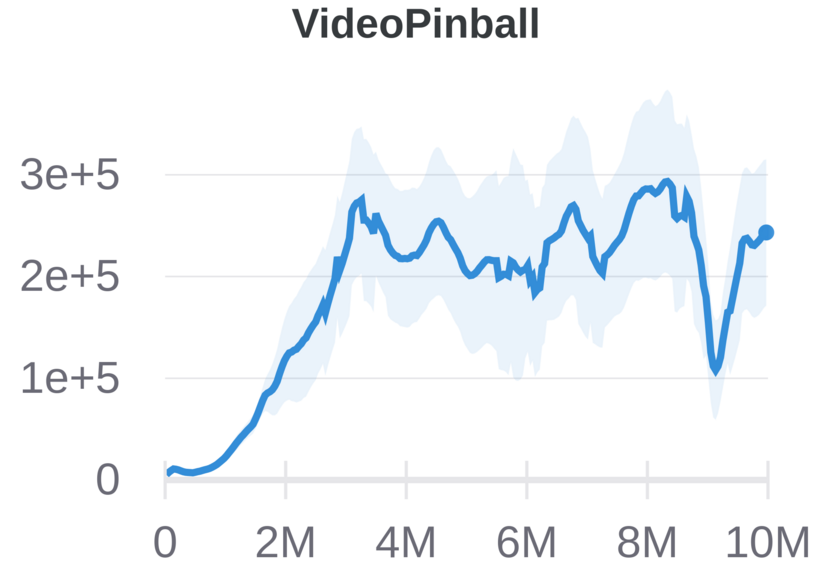} \\

    	\includegraphics[width=.18\linewidth,valign=m]{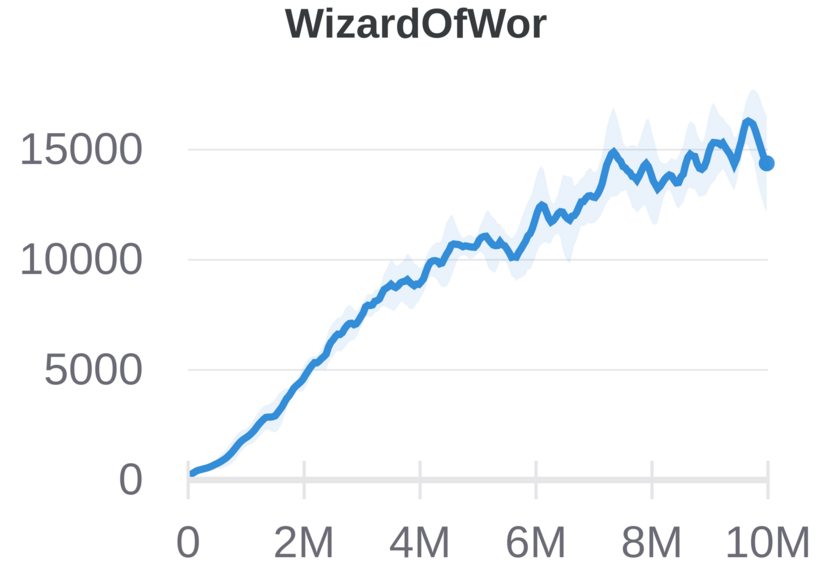} & 
    	\includegraphics[width=.18\linewidth,valign=m]{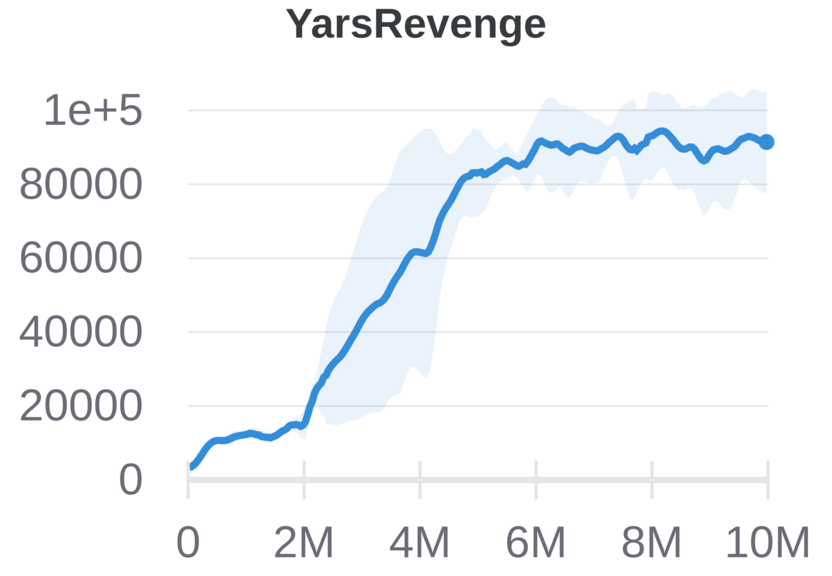} & 
    	\includegraphics[width=.18\linewidth,valign=m]{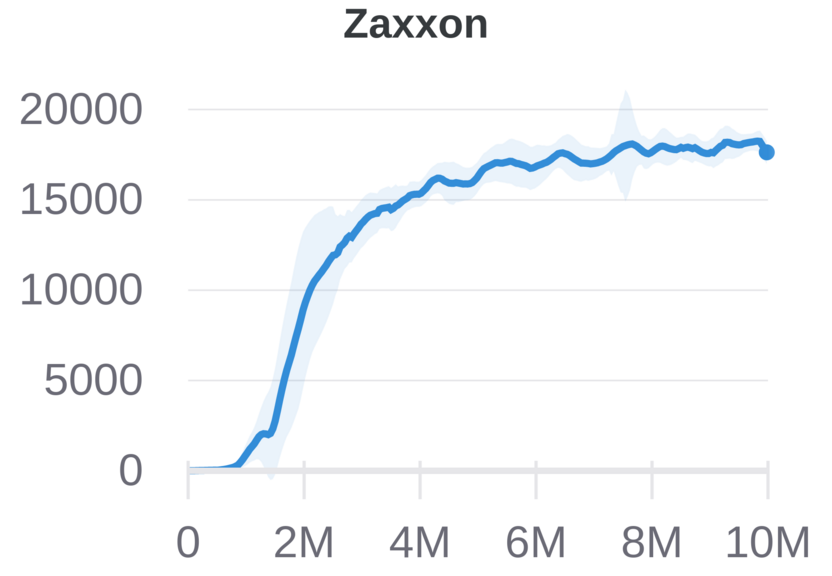} \\

    	\end{tabular}
    	\vspace*{7pt}
    	\caption{Learning curves for our Rainbow implementation for each of the selected Atari environments. Each curve represents the median 100-episode running average of episode returns over three seeds.}
    \end{figure}

\newpage

\begin{table}[!htp]\centering
    \renewcommand{\arraystretch}{1.2}
    \scriptsize
    \begin{tabular}{l >{\raggedleft}m{4.5em} >{\raggedleft}m{4.5em} >{\raggedleft}m{4.5em} >{\raggedleft}m{4.5em} >{\raggedleft}m{4.5em} >{\raggedleft}m{4.5em} >{\raggedleft\arraybackslash}m{4.5em}}
\textbf{} &\textbf{Uniform Random Policy} &\textbf{Human} &\textbf{Rainbow \citep{dopamine}} &\textbf{DQN \citep{mnih2013}} &\textbf{Rainbow \citep{dopamine}} &\textbf{Rainbow \citep{rainbow}} &\textbf{Ours} \\\midrule
\textbf{Alien} &\cellcolor[HTML]{fff2cc}227.8 &\cellcolor[HTML]{d3e4bb}6,875.4 &\cellcolor[HTML]{f9f0ca}1,241.6 &\cellcolor[HTML]{edecc5}3,069.3 &\cellcolor[HTML]{eaecc4}3,456.9 &\cellcolor[HTML]{c2deb4}9,491.7 &\cellcolor[HTML]{aed8ad}12,508.0 \\
\textbf{Amidar} &\cellcolor[HTML]{fff2cc}5.8 &\cellcolor[HTML]{c9e1b7}1,675.8 &\cellcolor[HTML]{f4efc8}348.3 &\cellcolor[HTML]{e7ebc3}739.5 &\cellcolor[HTML]{add7ac}2,529.1 &\cellcolor[HTML]{57bb8a}5,131.2 &\cellcolor[HTML]{bcdcb2}2,071.2 \\
\textbf{Assault} &\cellcolor[HTML]{fff2cc}222.4 &\cellcolor[HTML]{f0edc6}1,496.4 &\cellcolor[HTML]{f0eec7}1,483.1 &\cellcolor[HTML]{dae6be}3,358.6 &\cellcolor[HTML]{dbe7be}3,228.8 &\cellcolor[HTML]{57bb8a}14,198.5 &\cellcolor[HTML]{81c99b}10,709.5 \\
\textbf{Asterix} &\cellcolor[HTML]{fff2cc}210.0 &\cellcolor[HTML]{fcf1cb}8,503.3 &\cellcolor[HTML]{fef2cc}2,862.7 &\cellcolor[HTML]{fdf2cc}6,011.6 &\cellcolor[HTML]{f8f0ca}18,366.6 &\cellcolor[HTML]{57bb8a}428,200.3 &\cellcolor[HTML]{77c697}346,758.2 \\
\textbf{Asteroids} &\cellcolor[HTML]{fff2cc}719.1 &\cellcolor[HTML]{57bb8a}13,156.7 &\cellcolor[HTML]{fef2cc}809.9 &\cellcolor[HTML]{f3eec8}1,629.3 &\cellcolor[HTML]{f5efc8}1,483.5 &\cellcolor[HTML]{e5eac2}2,712.8 &\cellcolor[HTML]{62bf8f}12,345.9 \\
\textbf{Atlantis} &\cellcolor[HTML]{fff2cc}12,850.0 &\cellcolor[HTML]{fcf1cb}29,028.1 &\cellcolor[HTML]{dfe8c0}170,363.8 &\cellcolor[HTML]{f0eec7}85,950.0 &\cellcolor[HTML]{5cbd8c}802,548.0 &\cellcolor[HTML]{57bb8a}826,659.5 &\cellcolor[HTML]{5abc8c}812,825.9 \\
\textbf{BankHeist} &\cellcolor[HTML]{fff2cc}14.2 &\cellcolor[HTML]{a9d6aa}734.4 &\cellcolor[HTML]{94cfa2}904.5 &\cellcolor[HTML]{cee2b9}429.7 &\cellcolor[HTML]{80c99a}1,075.0 &\cellcolor[HTML]{5ebe8d}1,358.0 &\cellcolor[HTML]{57bb8a}1,411.1 \\
\textbf{BattleZone} &\cellcolor[HTML]{fff2cc}2,360.0 &\cellcolor[HTML]{cae1b7}37,800.0 &\cellcolor[HTML]{e1e8c0}22,448.7 &\cellcolor[HTML]{dbe7be}26,300.0 &\cellcolor[HTML]{c6e0b6}40,060.6 &\cellcolor[HTML]{a5d5a9}62,010.0 &\cellcolor[HTML]{57bb8a}112,652.3 \\
\textbf{BeamRider} &\cellcolor[HTML]{fff2cc}363.9 &\cellcolor[HTML]{dde7bf}5,774.7 &\cellcolor[HTML]{dfe8c0}5,434.9 &\cellcolor[HTML]{d6e5bc}6,845.9 &\cellcolor[HTML]{d9e6bd}6,290.5 &\cellcolor[HTML]{95d0a3}16,850.2 &\cellcolor[HTML]{57bb8a}26,398.8 \\
\textbf{Berzerk} &\cellcolor[HTML]{fff2cc}123.7 &\cellcolor[HTML]{7ec89a}2,630.4 &\cellcolor[HTML]{eeedc6}456.1 &\cellcolor[HTML]{e8ebc3}585.6 &\cellcolor[HTML]{dbe7be}833.4 &\cellcolor[HTML]{83ca9c}2,545.6 &\cellcolor[HTML]{57bb8a}3,388.0 \\
\textbf{Bowling} &\cellcolor[HTML]{fff2cc}23.1 &\cellcolor[HTML]{57bb8a}154.8 &\cellcolor[HTML]{e5eac2}44.1 &\cellcolor[HTML]{e7eac3}42.4 &\cellcolor[HTML]{e6eac3}42.9 &\cellcolor[HTML]{f7f0c9}30.0 &\cellcolor[HTML]{e9ebc4}40.9 \\
\textbf{Boxing} &\cellcolor[HTML]{fff2cc}0.1 &\cellcolor[HTML]{f8f0ca}4.3 &\cellcolor[HTML]{7fc89a}76.3 &\cellcolor[HTML]{87cb9d}71.8 &\cellcolor[HTML]{59bc8b}98.6 &\cellcolor[HTML]{58bc8b}99.6 &\cellcolor[HTML]{57bb8a}99.7 \\
\textbf{Breakout} &\cellcolor[HTML]{fff2cc}1.7 &\cellcolor[HTML]{f6efc9}31.8 &\cellcolor[HTML]{f0eec7}49.9 &\cellcolor[HTML]{82c99b}401.2 &\cellcolor[HTML]{dae6be}120.1 &\cellcolor[HTML]{7dc899}417.5 &\cellcolor[HTML]{57bb8a}537.6 \\
\textbf{Centipede} &\cellcolor[HTML]{fff2cc}2,090.9 &\cellcolor[HTML]{57bb8a}11,963.2 &\cellcolor[HTML]{d0e3ba}4,877.8 &\cellcolor[HTML]{96d0a3}8,309.4 &\cellcolor[HTML]{b4daaf}6,509.9 &\cellcolor[HTML]{98d1a4}8,167.3 &\cellcolor[HTML]{95d0a3}8,368.2 \\
\textbf{ChopperCommand} &\cellcolor[HTML]{fff2cc}811.0 &\cellcolor[HTML]{9fd3a7}9,881.8 &\cellcolor[HTML]{eeedc6}2,447.4 &\cellcolor[HTML]{c1deb4}6,686.7 &\cellcolor[HTML]{85ca9c}12,337.5 &\cellcolor[HTML]{57bb8a}16,654.0 &\cellcolor[HTML]{dbe7be}4,208.0 \\
\textbf{CrazyClimber} &\cellcolor[HTML]{f8f0c9}10,780.5 &\cellcolor[HTML]{dfe8c0}35,410.5 &\cellcolor[HTML]{96d0a3}107,805.4 &\cellcolor[HTML]{90cea1}114,103.3 &\cellcolor[HTML]{70c494}145,389.3 &\cellcolor[HTML]{59bc8b}168,788.5 &\cellcolor[HTML]{75c596}140,712.0 \\
\textbf{Defender} &\cellcolor[HTML]{fff2cc}2,874.5 &\cellcolor[HTML]{f0edc6}18,688.9 & &\cellcolor[HTML]{ebecc4}23,633.0 & &\cellcolor[HTML]{cbe1b8}55,105.0 &\cellcolor[HTML]{57bb8a}169,929.6 \\
\textbf{DemonAttack} &\cellcolor[HTML]{fff2cc}152.1 &\cellcolor[HTML]{fbf1cb}3,401.3 &\cellcolor[HTML]{fcf1cb}2,943.1 &\cellcolor[HTML]{f3efc8}9,711.2 &\cellcolor[HTML]{eaebc4}17,071.3 &\cellcolor[HTML]{72c495}111,185.2 &\cellcolor[HTML]{57bb8a}131,657.2 \\
\textbf{DoubleDunk} &\cellcolor[HTML]{fff2cc}-18.6 &\cellcolor[HTML]{f3eec7}-15.5 &\cellcolor[HTML]{fff2cc}-18.6 &\cellcolor[HTML]{fdf2cc}-18.1 &\cellcolor[HTML]{57bb8a}22.1 &\cellcolor[HTML]{b4daaf}-0.3 &\cellcolor[HTML]{b8dbb0}-1.2 \\
\textbf{Enduro} &\cellcolor[HTML]{fff2cc}0.0 &\cellcolor[HTML]{e9ebc3}309.6 &\cellcolor[HTML]{83ca9c}1,680.3 &\cellcolor[HTML]{e9ebc4}301.8 &\cellcolor[HTML]{5cbd8c}2,200.2 &\cellcolor[HTML]{62bf8f}2,125.9 &\cellcolor[HTML]{57bb8a}2,266.0 \\
\textbf{FishingDerby} &\cellcolor[HTML]{fff2cc}-91.7 &\cellcolor[HTML]{85cb9d}5.5 &\cellcolor[HTML]{7ac798}14.5 &\cellcolor[HTML]{8dcda0}-0.8 &\cellcolor[HTML]{58bc8b}41.8 &\cellcolor[HTML]{65c090}31.3 &\cellcolor[HTML]{57bb8a}42.1 \\
\textbf{Freeway} &\cellcolor[HTML]{fff2cc}0.0 &\cellcolor[HTML]{6dc393}29.6 &\cellcolor[HTML]{61be8e}32.1 &\cellcolor[HTML]{6ac192}30.3 &\cellcolor[HTML]{59bc8b}33.7 &\cellcolor[HTML]{57bb8a}34.0 &\cellcolor[HTML]{58bc8b}34.0 \\
\textbf{Frostbite} &\cellcolor[HTML]{fff2cc}65.2 &\cellcolor[HTML]{b4daaf}4,334.7 &\cellcolor[HTML]{d2e4bb}2,647.0 &\cellcolor[HTML]{fbf1cb}328.3 &\cellcolor[HTML]{70c394}8,207.7 &\cellcolor[HTML]{57bb8a}9,590.5 &\cellcolor[HTML]{a3d4a8}5,282.1 \\
\textbf{Gopher} &\cellcolor[HTML]{fff2cc}257.6 &\cellcolor[HTML]{fbf1cb}2,321.0 &\cellcolor[HTML]{f6efc9}4,399.6 &\cellcolor[HTML]{ebecc4}8,777.4 &\cellcolor[HTML]{e7eac3}10,641.1 &\cellcolor[HTML]{57bb8a}70,354.6 &\cellcolor[HTML]{57bb8a}25,606.8 \\
\textbf{Gravitar} &\cellcolor[HTML]{fff2cc}173.0 &\cellcolor[HTML]{57bb8a}2,672.0 &\cellcolor[HTML]{fdf2cc}208.2 &\cellcolor[HTML]{f7f0c9}306.7 &\cellcolor[HTML]{b6daaf}1,271.8 &\cellcolor[HTML]{acd7ac}1,419.3 &\cellcolor[HTML]{7dc899}2,107.3 \\
\textbf{Hero} &\cellcolor[HTML]{fff2cc}1,027.0 &\cellcolor[HTML]{b4daaf}25,762.5 &\cellcolor[HTML]{e3e9c1}10,468.5 &\cellcolor[HTML]{c6e0b6}19,950.3 &\cellcolor[HTML]{74c596}46,675.2 &\cellcolor[HTML]{57bb8a}55,887.4 &\cellcolor[HTML]{d4e4bb}15,377.2 \\
\textbf{IceHockey} &\cellcolor[HTML]{fff2cc}-11.2 &\cellcolor[HTML]{8dcda0}0.9 &\cellcolor[HTML]{c6e0b6}-5.2 &\cellcolor[HTML]{a5d5a9}-1.6 &\cellcolor[HTML]{97d0a4}-0.2 &\cellcolor[HTML]{8bcc9f}1.1 &\cellcolor[HTML]{57bb8a}6.6 \\
\textbf{Kangaroo} &\cellcolor[HTML]{fff2cc}52.0 &\cellcolor[HTML]{dde7bf}3,035.0 &\cellcolor[HTML]{c0deb3}5,579.6 &\cellcolor[HTML]{b2d9ae}6,740.0 &\cellcolor[HTML]{6dc393}12,748.3 &\cellcolor[HTML]{57bb8a}14,637.5 &\cellcolor[HTML]{7cc799}11,498.7 \\
\textbf{Krull} &\cellcolor[HTML]{fff2cc}1,598.0 &\cellcolor[HTML]{f0edc6}2,394.6 &\cellcolor[HTML]{abd7ab}5,980.4 &\cellcolor[HTML]{d5e5bc}3,804.7 &\cellcolor[HTML]{d0e3ba}4,066.0 &\cellcolor[HTML]{76c596}8,741.5 &\cellcolor[HTML]{57bb8a}10,324.3 \\
\textbf{KungFuMaster} &\cellcolor[HTML]{fff2cc}258.5 &\cellcolor[HTML]{b6dbb0}22,736.2 &\cellcolor[HTML]{c2deb4}19,195.5 &\cellcolor[HTML]{b5daaf}23,270.0 &\cellcolor[HTML]{aad7ab}26,475.1 &\cellcolor[HTML]{57bb8a}52,181.0 &\cellcolor[HTML]{a7d6aa}27,444.4 \\
\textbf{MontezumaRevenge} &\cellcolor[HTML]{fff2cc}0.0 &\cellcolor[HTML]{f1eec7}4,367.0 &\cellcolor[HTML]{fff2cc}0.8 &\cellcolor[HTML]{fff2cc}0.0 &\cellcolor[HTML]{fef2cc}500.0 &\cellcolor[HTML]{fef2cc}384.0 &\cellcolor[HTML]{fff2cc}0.0 \\
\textbf{MsPacman} &\cellcolor[HTML]{fff2cc}307.3 &\cellcolor[HTML]{57bb8a}15,693.4 &\cellcolor[HTML]{e9ebc4}2,399.3 &\cellcolor[HTML]{eaebc4}2,311.0 &\cellcolor[HTML]{d9e6bd}3,861.0 &\cellcolor[HTML]{c8e0b7}5,380.4 &\cellcolor[HTML]{c2deb4}5,981.7 \\
\textbf{NameThisGame} &\cellcolor[HTML]{fff2cc}2,292.3 &\cellcolor[HTML]{eeedc6}4,076.2 &\cellcolor[HTML]{bfddb3}9,023.2 &\cellcolor[HTML]{d0e3ba}7,256.7 &\cellcolor[HTML]{bfddb3}9,025.8 &\cellcolor[HTML]{98d0a4}13,136.0 &\cellcolor[HTML]{57bb8a}19,819.0 \\
\textbf{Phoenix} &\cellcolor[HTML]{fff2cc}761.4 &\cellcolor[HTML]{f5efc9}7,242.6 &\cellcolor[HTML]{f9f0ca}4,925.3 &\cellcolor[HTML]{f3efc8}8,485.2 &\cellcolor[HTML]{f3efc8}8,545.4 &\cellcolor[HTML]{57bb8a}108,528.6 &\cellcolor[HTML]{a2d4a8}60,954.5 \\
\textbf{Pitfall} &\cellcolor[HTML]{fef2cc}-229.4 &\cellcolor[HTML]{57bb8a}6,463.7 &\cellcolor[HTML]{f8f0ca}-1.4 &\cellcolor[HTML]{fff2cc}-286.1 &\cellcolor[HTML]{f9f0ca}-19.8 &\cellcolor[HTML]{f8f0ca}0.0 &\cellcolor[HTML]{f8f0ca}-1.8 \\
\textbf{Pong} &\cellcolor[HTML]{fff2cc}-20.7 &\cellcolor[HTML]{87cb9d}9.3 &\cellcolor[HTML]{6dc293}15.8 &\cellcolor[HTML]{60be8e}18.9 &\cellcolor[HTML]{5bbd8c}20.2 &\cellcolor[HTML]{58bc8b}20.9 &\cellcolor[HTML]{57bb8a}21.0 \\
\textbf{PrivateEye} &\cellcolor[HTML]{fff2cc}24.9 &\cellcolor[HTML]{57bb8a}69,571.3 &\cellcolor[HTML]{fff2cc}89.2 &\cellcolor[HTML]{fbf1cb}1,787.6 &\cellcolor[HTML]{cce2b8}21,333.6 &\cellcolor[HTML]{f5efc9}4,234.0 &\cellcolor[HTML]{fff2cc}253.8 \\
\textbf{Qbert} &\cellcolor[HTML]{fff2cc}163.9 &\cellcolor[HTML]{bdddb2}13,455.0 &\cellcolor[HTML]{dee8bf}6,861.5 &\cellcolor[HTML]{cbe1b8}10,595.8 &\cellcolor[HTML]{aad6ab}17,382.9 &\cellcolor[HTML]{57bb8a}33,817.5 &\cellcolor[HTML]{80c99a}25,712.4 \\
\textbf{RoadRunner} &\cellcolor[HTML]{fff2cc}11.5 &\cellcolor[HTML]{efedc6}7,845.0 &\cellcolor[HTML]{b9dbb1}34,454.9 &\cellcolor[HTML]{dae6be}18,256.7 &\cellcolor[HTML]{8fcea0}54,662.1 &\cellcolor[HTML]{80c99a}62,041.0 &\cellcolor[HTML]{57bb8a}81,831.7 \\
\textbf{Robotank} &\cellcolor[HTML]{fff2cc}2.2 &\cellcolor[HTML]{e8ebc3}11.9 &\cellcolor[HTML]{d0e3ba}21.7 &\cellcolor[HTML]{86cb9d}51.6 &\cellcolor[HTML]{64c08f}65.5 &\cellcolor[HTML]{6ec393}61.4 &\cellcolor[HTML]{57bb8a}70.7 \\
\textbf{Seaquest} &\cellcolor[HTML]{fff2cc}68.4 &\cellcolor[HTML]{cae1b8}20,181.8 &\cellcolor[HTML]{fbf1cb}1,646.0 &\cellcolor[HTML]{f2eec7}5,286.0 &\cellcolor[HTML]{e6eac2}9,903.4 &\cellcolor[HTML]{d6e5bc}15,898.9 &\cellcolor[HTML]{57bb8a}63,724.4 \\
\textbf{Skiing} &\cellcolor[HTML]{afd8ad}-17,098.1 &\cellcolor[HTML]{57bb8a}-4,336.9 &\cellcolor[HTML]{dee8bf}-23,886.9 &\cellcolor[HTML]{94cfa2}-13,062.3 &\cellcolor[HTML]{fff2cc}-28,707.6 &\cellcolor[HTML]{93cfa2}-12,957.8 &\cellcolor[HTML]{d2e4bb}-22,076.8 \\
\textbf{Solaris} &\cellcolor[HTML]{fff2cc}1,236.3 &\cellcolor[HTML]{57bb8a}12,326.7 &\cellcolor[HTML]{fdf2cb}1,429.0 &\cellcolor[HTML]{dde7bf}3,482.8 &\cellcolor[HTML]{faf1ca}1,582.7 &\cellcolor[HTML]{dce7bf}3,560.3 &\cellcolor[HTML]{e7eac3}2,877.6 \\
\textbf{SpaceInvaders} &\cellcolor[HTML]{fff2cc}148.0 &\cellcolor[HTML]{f6f0c9}1,652.3 &\cellcolor[HTML]{fcf1cb}769.0 &\cellcolor[HTML]{f5efc8}1,975.5 &\cellcolor[HTML]{e8ebc3}4,130.9 &\cellcolor[HTML]{8fcea0}18,789.0 &\cellcolor[HTML]{57bb8a}28,098.6 \\
\textbf{StarGunner} &\cellcolor[HTML]{fff2cc}664.0 &\cellcolor[HTML]{faf1ca}10,250.0 &\cellcolor[HTML]{fff2cc}1,536.9 &\cellcolor[HTML]{e0e8c0}57,996.7 &\cellcolor[HTML]{e0e8c0}57,908.7 &\cellcolor[HTML]{bbdcb2}127,029.0 &\cellcolor[HTML]{57bb8a}310,403.7 \\
\textbf{Tennis} &\cellcolor[HTML]{fff2cc}-23.8 &\cellcolor[HTML]{c0deb4}-8.9 &\cellcolor[HTML]{a5d5a9}-2.3 &\cellcolor[HTML]{a5d5a9}-2.5 &\cellcolor[HTML]{9cd2a5}-0.2 &\cellcolor[HTML]{9bd2a5}0.0 &\cellcolor[HTML]{57bb8a}15.9 \\
\textbf{TimePilot} &\cellcolor[HTML]{fcf1cb}3,568.0 &\cellcolor[HTML]{eeedc6}5,925.0 &\cellcolor[HTML]{fff2cc}2,960.6 &\cellcolor[HTML]{eeedc6}5,946.7 &\cellcolor[HTML]{cae1b7}12,050.5 &\cellcolor[HTML]{c4dfb5}12,926.0 &\cellcolor[HTML]{57bb8a}31,333.2 \\
\textbf{Tutankham} &\cellcolor[HTML]{fff2cc}11.4 &\cellcolor[HTML]{8dcda0}167.6 &\cellcolor[HTML]{73c495}203.8 &\cellcolor[HTML]{7fc99a}186.7 &\cellcolor[HTML]{59bc8b}239.1 &\cellcolor[HTML]{57bb8a}241.0 &\cellcolor[HTML]{8ecda0}167.0 \\
\textbf{Venture} &\cellcolor[HTML]{fff2cc}0.0 &\cellcolor[HTML]{7dc899}1,187.5 &\cellcolor[HTML]{fef2cc}17.2 &\cellcolor[HTML]{d6e5bc}380.0 &\cellcolor[HTML]{57bb8a}1,528.9 &\cellcolor[HTML]{fff2cc}5.5 &\cellcolor[HTML]{cfe3ba}437.1 \\
\textbf{VideoPinball} &\cellcolor[HTML]{fff2cc}16,256.9 &\cellcolor[HTML]{fff2cc}17,297.6 &\cellcolor[HTML]{fbf1cb}30,298.4 &\cellcolor[HTML]{f7f0c9}42,684.1 &\cellcolor[HTML]{6dc393}466,895.0 &\cellcolor[HTML]{57bb8a}533,936.5 &\cellcolor[HTML]{add8ac}269,619.0 \\
\textbf{WizardOfWor} &\cellcolor[HTML]{fff2cc}563.5 &\cellcolor[HTML]{d7e5bd}4,756.5 &\cellcolor[HTML]{eaecc4}2,727.9 &\cellcolor[HTML]{e4eac2}3,393.3 &\cellcolor[HTML]{b8dbb1}7,878.6 &\cellcolor[HTML]{57bb8a}17,862.5 &\cellcolor[HTML]{6ec393}15,518.6 \\
\textbf{YarsRevenge} &\cellcolor[HTML]{fff2cc}3,092.9 &\cellcolor[HTML]{a9d6aa}54,576.9 &\cellcolor[HTML]{f3eec8}10,536.7 &\cellcolor[HTML]{e6eac3}18,089.9 &\cellcolor[HTML]{b8dbb0}45,542.0 &\cellcolor[HTML]{57bb8a}102,557.0 &\cellcolor[HTML]{5ebe8d}98,908.0 \\
\textbf{Zaxxon} &\cellcolor[HTML]{fff2cc}32.5 &\cellcolor[HTML]{badcb1}9,173.3 &\cellcolor[HTML]{dde7bf}4,521.1 &\cellcolor[HTML]{dae6be}4,976.7 &\cellcolor[HTML]{91cea1}14,603.0 &\cellcolor[HTML]{57bb8a}22,209.5 &\cellcolor[HTML]{71c495}18,832.6 \\
& & & & & & & \\
\textbf{training frames} & & &\cellcolor[HTML]{c9daf8}10M &\cellcolor[HTML]{ead1dc}200M &\cellcolor[HTML]{ead1dc}200M &\cellcolor[HTML]{ead1dc}200M &\cellcolor[HTML]{c9daf8}10M \\
\textbf{mean HNS} &\cellcolor[HTML]{fff2cc}0 &\cellcolor[HTML]{f5efc8}100 &\cellcolor[HTML]{eeedc6}166.9 &\cellcolor[HTML]{e8ebc3}224.2 &\cellcolor[HTML]{87cb9d}1161.6 &\cellcolor[HTML]{57bb8a}1624.9 &\cellcolor[HTML]{86cb9d}1174.2 \\
\textbf{median HNS} &\cellcolor[HTML]{fff2cc}0 &\cellcolor[HTML]{b7dbb0}100 &\cellcolor[HTML]{dbe7be}50.4 &\cellcolor[HTML]{c6e0b6}79.3 &\cellcolor[HTML]{95d0a3}145.8 &\cellcolor[HTML]{57bb8a}231.0 &\cellcolor[HTML]{6ac292}205.7 \\
\textbf{\# games above human} &\cellcolor[HTML]{fff2cc}0 &\cellcolor[HTML]{fff2cc}0 &\cellcolor[HTML]{b0d8ad}19 &\cellcolor[HTML]{9bd1a5}24 &\cellcolor[HTML]{60be8e}38 &\cellcolor[HTML]{57bb8a}40 &\cellcolor[HTML]{5cbd8c}39 \\
    \end{tabular}
    \vspace*{10pt}
    \caption{Evaluation scores for all 53 tested Atari games (averaged over 3 random seeds). Random and human scores are from \citet{mnih2015} where available, otherwise from \citet{agent57}.}
\end{table}

\section{Hyperparameters}
\label{appendix:hyp}

\subsection{Hyperparameters for Rainbow}

This section lists the hyperparameters used in our experiments. Parameters that differ from the ones used in \citet{rainbow} are marked with an asterisk. As in previous work, the unit "frames" refers to the number of environment steps taken by the wrapped environment, including frame-skipping. For noisy-nets DQN we implemented the "factorized Gaussian noise" variant, with noise vectors generated on the GPU.

\begin{center}
\begin{tabular}{ lc } 
 \textbf{Parameter} & \textbf{Value} \\
 \hline
 
 Discount factor $\gamma$ & 0.99 \\ 
 Q-target update frequency & 32,000 frames \\  
 Importance sampling $\beta_0$ for PER & 0.45 \\
 $n$ in n-step bootstrapping & 3 \\
 Initial exploration $\epsilon$ & 1.0 \\
 Final exploration $\epsilon$ & 0.01 \\
 *Exploration $\epsilon$ decay time & 500,000 frames \\
 $\sigma_0$ for noisy linear layers & 0.5 \\
 
 \hline
 
 *Learning rate & 0.00025 \\
 *Adam $\epsilon$ parameter & 0.005$/$batch size\\
 Gradient clip norm & 10 \\
 Loss function & Huber \\
 
 \hline
 
 *Batch size & 256 \\ 
 *Parallel environments & 64 \\ 
 Replay Buffer Size & ~1M transitions \\ 
 Training starts at & 80,000 frames \\ 
 *Q-network architecture & IMPALA-large with 2x channels \\
\end{tabular}
\end{center}

\subsection{Environment pre-processing hyperparameters}

This section lists the settings for preprocessing environments from \texttt{gym}, \texttt{gym-retro} and \texttt{procgen}. All environments used a time limit of 108k frames (30 minutes of emulator time). Image downscaling was performed with area interpolation. For \texttt{gym} environments, we max-pooled consecutive frames and used 0-30 noop actions at the beginning of each episode as in \citet{rainbow}.

\begin{center}
\begin{tabular}{ llc } 
 \textbf{Parameter} & \textbf{Environment} & \textbf{Value} \\
 \hline
 
  & gym & yes \\ 
  Grayscale & retro & no \\ 
  & procgen & no \\ 
  
 \hline
 
   & gym & 4 \\ 
  Frame-skipping & retro & 4 \\ 
  & procgen & 1 \\ 
  
   \hline
  
   & gym & 4 \\ 
  Frame-stacking & retro & 4 \\ 
  & procgen & 4 \\ 
 
  \hline
  
  & gym & 84 $\times$ 84 \\ 
  Resolution & retro & 72 $\times$ 96 \\ 
  & procgen & 64 $\times$ 64 \\
  
\end{tabular}
\end{center}

\newpage
\subsection{Network Architecture}

\begin{figure}[h]
	\begin{center}
		\includegraphics[width=0.6\columnwidth]{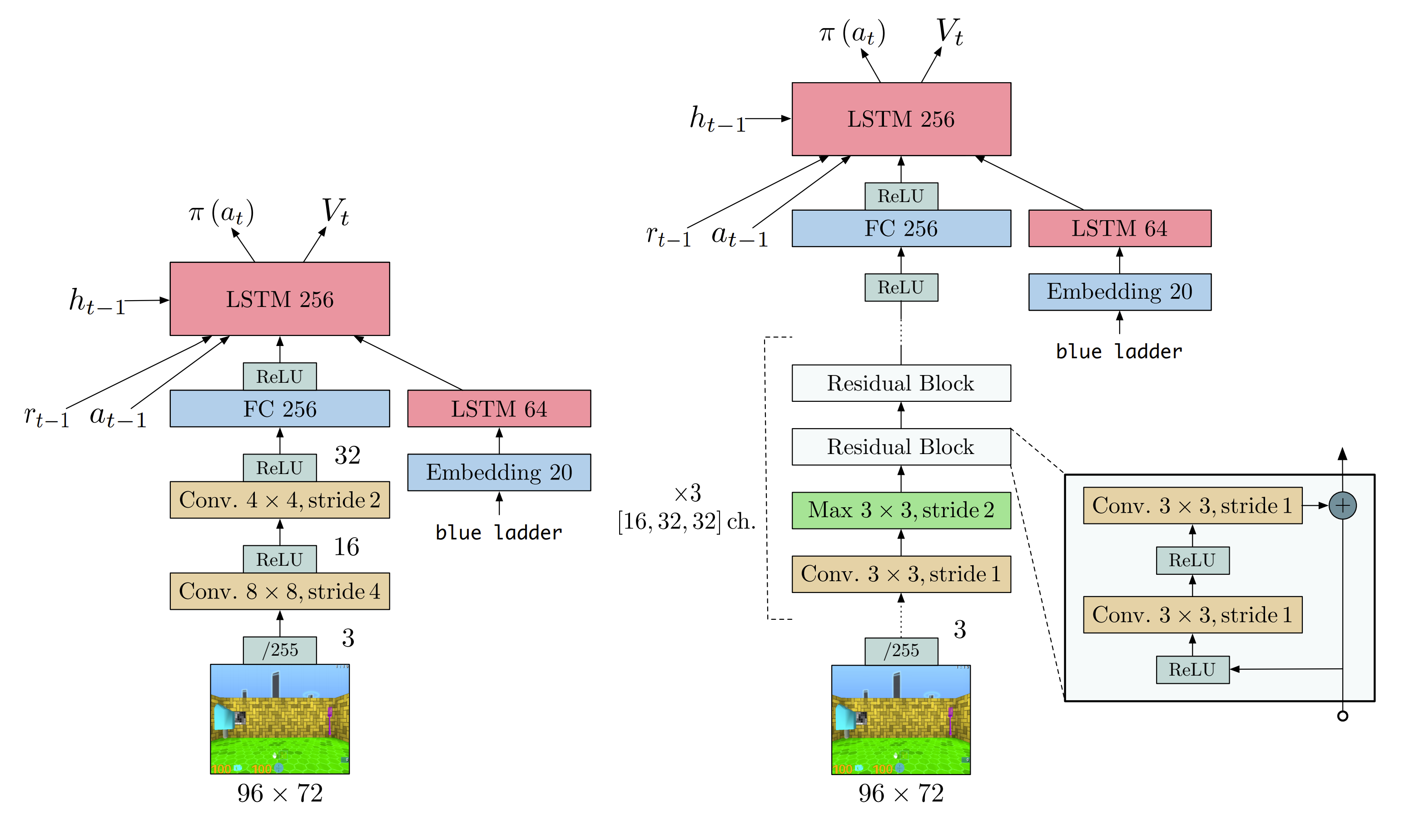}
	\end{center}\caption{The unmodified small (left) and large (right) IMPALA CNN network architecture \citep[from][]{impala}.}
	\label{fig:impala}
\end{figure}

\end{document}